\documentclass[english]{svjour3}
\usepackage[T1]{fontenc}
\usepackage[latin9]{inputenc}
\usepackage{babel}
\usepackage{array}
\usepackage{float}
\usepackage{wrapfig}
\usepackage{url}
\usepackage{multirow}
\usepackage{amsmath}
\usepackage{amssymb}
\usepackage{graphicx}
\usepackage[authoryear]{natbib}
\usepackage{xargs}[2008/03/08]
\PassOptionsToPackage{normalem}{ulem}
\usepackage{ulem}
\usepackage[unicode=true,pdfusetitle,
 bookmarks=true,bookmarksnumbered=false,bookmarksopen=false,
 breaklinks=false,pdfborder={0 0 1},backref=false,colorlinks=false]
 {hyperref}

\makeatletter

\providecommand{\tabularnewline}{\\}
\floatstyle{ruled}
\newfloat{algorithm}{tbp}{loa}
\providecommand{\algorithmname}{Algorithm}
\floatname{algorithm}{\protect\algorithmname}

\usepackage{algolyx}
\usepackage{graphicx}
\usepackage{mathptmx}

\makeatother

\begin{document}

\title{Scalable Semi-supervised Learning with Graph-based Kernel Machine}

\author{Trung Le \and  Khanh Nguyen \and  Van Nguyen \and  Vu Nguyen \and 
Dinh Phung}

\institute{Trung Le, Khanh Nguyen, Van Nguyen, Vu Nguyen, Dinh Phung \at Centre
for Pattern Recognition and Data Analytics, Deakin University, Australia\\
Tel.: +61-352479655\\
\email{\{trung.l, khanhnk, van.n, v.nguyen, dinh.phung\}@deakin.edu.au}\\
}

\date{Received: date / Accepted: date\newcommand{\sidenote}[1]{\marginpar{\small \emph{\color{Medium}#1}}}

\global\long\def\se{\hat{\text{se}}}

\global\long\def\interior{\text{int}}

\global\long\def\boundary{\text{bd}}

\global\long\def\ML{\textsf{ML}}

\global\long\def\GML{\mathsf{GML}}

\global\long\def\HMM{\mathsf{HMM}}

\global\long\def\support{\text{supp}}

\global\long\def\new{\text{*}}

\global\long\def\stir{\text{Stirl}}

\global\long\def\mA{\mathcal{A}}

\global\long\def\mB{\mathcal{B}}

\global\long\def\mF{\mathcal{F}}

\global\long\def\mK{\mathcal{K}}

\global\long\def\mH{\mathcal{H}}

\global\long\def\mX{\mathcal{X}}

\global\long\def\mZ{\mathcal{Z}}

\global\long\def\mS{\mathcal{S}}

\global\long\def\Ical{\mathcal{I}}

\global\long\def\mT{\mathcal{T}}

\global\long\def\Pcal{\mathcal{P}}

\global\long\def\dist{d}

\global\long\def\HX{\entro\left(X\right)}
 \global\long\def\entropyX{\HX}

\global\long\def\HY{\entro\left(Y\right)}
 \global\long\def\entropyY{\HY}

\global\long\def\HXY{\entro\left(X,Y\right)}
 \global\long\def\entropyXY{\HXY}

\global\long\def\mutualXY{\mutual\left(X;Y\right)}
 \global\long\def\mutinfoXY{\mutualXY}

\global\long\def\given{\mid}

\global\long\def\gv{\given}

\global\long\def\goto{\rightarrow}

\global\long\def\asgoto{\stackrel{a.s.}{\longrightarrow}}

\global\long\def\pgoto{\stackrel{p}{\longrightarrow}}

\global\long\def\dgoto{\stackrel{d}{\longrightarrow}}

\global\long\def\lik{\mathcal{L}}

\global\long\def\logll{\mathit{l}}

\global\long\def\vectorize#1{\mathbf{#1}}

\global\long\def\vt#1{\mathbf{#1}}

\global\long\def\gvt#1{\boldsymbol{#1}}

\global\long\def\idp{\ \bot\negthickspace\negthickspace\bot\ }
 \global\long\def\cdp{\idp}

\global\long\def\das{}

\global\long\def\id{\mathbb{I}}

\global\long\def\idarg#1#2{\id\left\{  #1,#2\right\}  }

\global\long\def\iid{\stackrel{\text{iid}}{\sim}}

\global\long\def\bzero{\vt 0}

\global\long\def\bone{\mathbf{1}}

\global\long\def\boldm{\boldsymbol{m}}

\global\long\def\bff{\vt f}

\global\long\def\bx{\boldsymbol{x}}

\global\long\def\bl{\boldsymbol{l}}

\global\long\def\bu{\boldsymbol{u}}

\global\long\def\bo{\boldsymbol{o}}

\global\long\def\bh{\boldsymbol{h}}

\global\long\def\bs{\boldsymbol{s}}

\global\long\def\bz{\boldsymbol{z}}

\global\long\def\xnew{y}

\global\long\def\bxnew{\boldsymbol{y}}

\global\long\def\bX{\boldsymbol{X}}

\global\long\def\tbx{\tilde{\bx}}

\global\long\def\by{\boldsymbol{y}}

\global\long\def\bY{\boldsymbol{Y}}

\global\long\def\bZ{\boldsymbol{Z}}

\global\long\def\bU{\boldsymbol{U}}

\global\long\def\bv{\boldsymbol{v}}

\global\long\def\bn{\boldsymbol{n}}

\global\long\def\bV{\boldsymbol{V}}

\global\long\def\bI{\boldsymbol{I}}

\global\long\def\bw{\vt w}

\global\long\def\balpha{\gvt{\alpha}}

\global\long\def\bbeta{\gvt{\beta}}

\global\long\def\bmu{\gvt{\mu}}

\global\long\def\btheta{\boldsymbol{\theta}}

\global\long\def\blambda{\boldsymbol{\lambda}}

\global\long\def\bgamma{\boldsymbol{\gamma}}

\global\long\def\bpsi{\boldsymbol{\psi}}

\global\long\def\bphi{\boldsymbol{\phi}}

\global\long\def\bpi{\boldsymbol{\pi}}

\global\long\def\bomega{\boldsymbol{\omega}}

\global\long\def\bepsilon{\boldsymbol{\epsilon}}

\global\long\def\btau{\boldsymbol{\tau}}

\global\long\def\bxi{\boldsymbol{\xi}}

\global\long\def\realset{\mathbb{R}}

\global\long\def\realn{\realset^{n}}

\global\long\def\integerset{\mathbb{Z}}

\global\long\def\natset{\integerset}

\global\long\def\integer{\integerset}

\global\long\def\natn{\natset^{n}}

\global\long\def\rational{\mathbb{Q}}

\global\long\def\rationaln{\rational^{n}}

\global\long\def\complexset{\mathbb{C}}

\global\long\def\comp{\complexset}

\global\long\def\compl#1{#1^{\text{c}}}

\global\long\def\and{\cap}

\global\long\def\compn{\comp^{n}}

\global\long\def\comb#1#2{\left({#1\atop #2}\right) }

\global\long\def\nchoosek#1#2{\left({#1\atop #2}\right)}

\global\long\def\param{\vt w}

\global\long\def\Param{\Theta}

\global\long\def\meanparam{\gvt{\mu}}

\global\long\def\Meanparam{\mathcal{M}}

\global\long\def\meanmap{\mathbf{m}}

\global\long\def\logpart{A}

\global\long\def\simplex{\Delta}

\global\long\def\simplexn{\simplex^{n}}

\global\long\def\dirproc{\text{DP}}

\global\long\def\ggproc{\text{GG}}

\global\long\def\DP{\text{DP}}

\global\long\def\ndp{\text{nDP}}

\global\long\def\hdp{\text{HDP}}

\global\long\def\gempdf{\text{GEM}}

\global\long\def\rfs{\text{RFS}}

\global\long\def\bernrfs{\text{BernoulliRFS}}

\global\long\def\poissrfs{\text{PoissonRFS}}

\global\long\def\grad{\gradient}
 \global\long\def\gradient{\nabla}

\global\long\def\partdev#1#2{\partialdev{#1}{#2}}
 \global\long\def\partialdev#1#2{\frac{\partial#1}{\partial#2}}

\global\long\def\partddev#1#2{\partialdevdev{#1}{#2}}
 \global\long\def\partialdevdev#1#2{\frac{\partial^{2}#1}{\partial#2\partial#2^{\top}}}

\global\long\def\closure{\text{cl}}

\global\long\def\cpr#1#2{\Pr\left(#1\ |\ #2\right)}

\global\long\def\var{\text{Var}}

\global\long\def\Var#1{\text{Var}\left[#1\right]}

\global\long\def\cov{\text{Cov}}

\global\long\def\Cov#1{\cov\left[ #1 \right]}

\global\long\def\COV#1#2{\underset{#2}{\cov}\left[ #1 \right]}

\global\long\def\corr{\text{Corr}}

\global\long\def\sst{\text{T}}

\global\long\def\SST{\sst}

\global\long\def\ess{\mathbb{E}}

\global\long\def\Ess#1{\ess\left[#1\right]}

\newcommandx\ESS[2][usedefault, addprefix=\global, 1=]{\underset{#2}{\ess}\left[#1\right]}

\global\long\def\fisher{\mathcal{F}}

\global\long\def\bfield{\mathcal{B}}
 \global\long\def\borel{\mathcal{B}}

\global\long\def\bernpdf{\text{Bernoulli}}

\global\long\def\betapdf{\text{Beta}}

\global\long\def\dirpdf{\text{Dir}}

\global\long\def\gammapdf{\text{Gamma}}

\global\long\def\gaussden#1#2{\text{Normal}\left(#1, #2 \right) }

\global\long\def\gauss{\mathbf{N}}

\global\long\def\gausspdf#1#2#3{\text{Normal}\left( #1 \lcabra{#2, #3}\right) }

\global\long\def\multpdf{\text{Mult}}

\global\long\def\poiss{\text{Pois}}

\global\long\def\poissonpdf{\text{Poisson}}

\global\long\def\pgpdf{\text{PG}}

\global\long\def\wshpdf{\text{Wish}}

\global\long\def\iwshpdf{\text{InvWish}}

\global\long\def\nwpdf{\text{NW}}

\global\long\def\niwpdf{\text{NIW}}

\global\long\def\studentpdf{\text{Student}}

\global\long\def\unipdf{\text{Uni}}

\global\long\def\transp#1{\transpose{#1}}
 \global\long\def\transpose#1{#1^{\mathsf{T}}}

\global\long\def\mgt{\succ}

\global\long\def\mge{\succeq}

\global\long\def\idenmat{\mathbf{I}}

\global\long\def\trace{\mathrm{tr}}

\global\long\def\argmax#1{\underset{_{#1}}{\text{argmax}} }

\global\long\def\argmin#1{\underset{_{#1}}{\text{argmin}\ } }

\global\long\def\diag{\text{diag}}

\global\long\def\norm{}

\global\long\def\spn{\text{span}}

\global\long\def\vtspace{\mathcal{V}}

\global\long\def\field{\mathcal{F}}
 \global\long\def\ffield{\mathcal{F}}

\global\long\def\inner#1#2{\left\langle #1,#2\right\rangle }
 \global\long\def\iprod#1#2{\inner{#1}{#2}}

\global\long\def\dprod#1#2{#1 \cdot#2}

\global\long\def\norm#1{\left\Vert #1\right\Vert }

\global\long\def\entro{\mathbb{H}}

\global\long\def\entropy{\mathbb{H}}

\global\long\def\Entro#1{\entro\left[#1\right]}

\global\long\def\Entropy#1{\Entro{#1}}

\global\long\def\mutinfo{\mathbb{I}}

\global\long\def\relH{\mathit{D}}

\global\long\def\reldiv#1#2{\relH\left(#1||#2\right)}

\global\long\def\KL{KL}

\global\long\def\KLdiv#1#2{\KL\left(#1\parallel#2\right)}
 \global\long\def\KLdivergence#1#2{\KL\left(#1\ \parallel\ #2\right)}

\global\long\def\crossH{\mathcal{C}}
 \global\long\def\crossentropy{\mathcal{C}}

\global\long\def\crossHxy#1#2{\crossentropy\left(#1\parallel#2\right)}

\global\long\def\breg{\text{BD}}

\global\long\def\lcabra#1{\left|#1\right.}

\global\long\def\lbra#1{\lcabra{#1}}

\global\long\def\rcabra#1{\left.#1\right|}

\global\long\def\rbra#1{\rcabra{#1}}

}
\maketitle
\begin{abstract}
Acquiring labels are often costly, whereas unlabeled data are usually
easy to obtain in modern machine learning applications. Semi-supervised
learning provides a principled machine learning framework to address
such situations, and has been applied successfully in many real-word
applications and industries. Nonetheless, most of existing semi-supervised
learning methods encounter two serious limitations when applied to
modern and large-scale datasets: computational burden and memory usage
demand. To this end, we present in this paper the \emph{Graph-based
semi-supervised Kernel Machine} (GKM), a method that leverages the
generalization ability of kernel-based method with the geometrical
and distributive information formulated through a spectral graph induced
from data for semi-supervised learning purpose. Our proposed GKM can
be solved directly in the primal form using the Stochastic Gradient
Descent method with the ideal convergence rate $\text{O}\left(\frac{1}{T}\right)$.
Besides, our formulation is suitable for a wide spectrum of important
loss functions in the literature of machine learning (i.e., Hinge,
smooth Hinge, Logistic, L1, and $\varepsilon$-insensitive) and smoothness
functions (i.e., $l_{p}\left(t\right)=\left|t\right|^{p}$ with $p\geq1$).
We further show that the well-known Laplacian Support Vector Machine
is a special case of our formulation. We validate our proposed method
on several benchmark datasets to demonstrate that GKM is appropriate
for the large-scale datasets since it is optimal in memory usage and
yields superior classification accuracy whilst simultaneously achieving
a significant computation speed-up in comparison with the state-of-the-art
baselines.  \vspace{-3mm}
\end{abstract}

\section{Introduction\label{sec:intro}}

Semi-supervised learning (SSL) aims at utilizing the intrinsic information
carried in unlabeled data to enhance the generalization capacity of
the learning algorithms. During the past decade, SSL has attracted
significant attention and has found applicable in a variety of real-world
problems including text categorization \cite{Joachims99}, image retrieval
\cite{WangCZ03}, bioinformatics \cite{kasabov2005}, natural language
processing \cite{Goutte2002} to name a few. While obtaining pre-defined
labels is a labor-intensive and time-consuming process \cite{Chapelle2008},
it is well known that unlabeled data, when being used in conjunction
with a small amount of labeled data, can bring a remarkable improvement
in classification accuracy \cite{Joachims99}.

A notable approach to semi-supervised learning paradigm is to employ
spectral graph in order to represent the adjacent and distributive
information carried in data. Graph-based methods are nonparametric,
discriminative, and transductive in nature. Typical graph-based methods
include min-cut \cite{blum2004}, harmonic function \cite{zhu03},
graph random walk \cite{Azran2007}, spectral graph transducer \cite{Joachims03transductivelearning,duong2015},
and manifold regularization \cite{Belkin2006}.

Inspired from the pioneering work of \cite{Joachims99}, recent works
have attempted to incorporate kernel methods such as Support Vector
Machine (SVM) \cite{svm_original} with the semi-supervised learning
paradigm. The underlying idea is to \textit{solve the standard SVM
problem while treating the unknown labels as optimization variables}
\cite{Chapelle2008}. This leads to a non-convex optimization problem
with a combinatorial explosion of label assignments. A wide spectrum
of techniques have been proposed to solve this non-convex optimization
problem, e.g., local combination search \cite{Joachims99}, gradient
descent \cite{Chapelle2005}, continuation techniques \cite{Chapelle2006},
convex-concave procedures \cite{Collobert06}, deterministic annealing
\cite{Sindhwani2006,le2013,nguyen2014}, and semi-definite programming
\cite{Bie06}. Although these works can somehow handle the combinatorial
intractability, their common requirement of repeatedly retraining
the model limits their applicability to real-world applications, hence
lacking the ability to perform online learning for large-scale applications.

Conjoining the advantages of kernel method and the spectral graph
theory, several existing works have tried to incorporate information
carried in a spectral graph for building a better kernel function
\cite{kondor2002diffusionkernel,Chapelle2003,SmolaKondor2003}. Basically,
these methods employ the Laplacian matrix induced from the spectral
graph to construct kernel functions which can capture the features
of the ambient space. Manifold regularization framework \cite{Belkin2006}
exploits the geometric information of the probability distribution
that generates data and incorporates it as an additional regularization
term. Two regularization terms are introduced to control the complexity
of the classifier in the ambient space and the complexity induced
from the geometric information of the distribution. However, the computational
complexity for manifold regularization approach is cubic in the training
size $n$ (i.e., $\textrm{O}\left(n^{3}\right)$). Hence other researches
have been carried out to enhance the scalability of the manifold regularization
framework \cite{Sindhwani05linearmanifold,TsangK06,Melacci2011}.
Specifically, the work of \cite{Melacci2011} makes use of the preconditioned
conjugate gradient to solve the optimization problem encountered in
manifold regularization framework in the primal form, reducing the
computational complexity from $\textrm{O}\left(n^{3}\right)$ to $\textrm{O}\left(n^{2}\right)$.
However, this approach is not suitable for online learning setting
since it actually solves the optimization problem in the first dual
layer instead of the primal form. In addition, the LapSVM in primal
approach \cite{Melacci2011} requires storing the entire Hessian matrix
of size $n\times n$ in the memory, resulting in a memory complexity
of $\text{O}(n^{2})$. Our evaluating experiments with LapSVM in primal
further show that it always consumes a huge amount of memory in its
execution (cf. Table \ref{tab:balanced80}). 

Recently, stochastic gradient descent (SGD) methods \cite{Shalev-shwartz07logarithmicregret,kakade2008duality,Lacoste2012}
have emerged as a promising framework to speed up the training process
and enable the online learning paradigm. SGD possesses three key advantages:
(1) it is fast; (2) it can be exploited to run in online mode; and
(3) it is efficient in memory usage. In this paper, we leverage the
strength from three bodies of theories, namely kernel method, spectral
graph theory and stochastic gradient descent to propose a novel approach
to semi-supervised learning, termed as \textit{Graph-based Semi-supervised
Kernel Machine} (GKM). Our GKM is applicable for a wide spectrum of
loss functions (cf. Section \ref{sec:lossfunc}) and smoothness functions
$l_{p}\left(.\right)$ where $p\geq1$ (cf. Eq. (\ref{eq:primal})).
In addition, we note that the well-known Laplacian Support Vector
Machine (LapSVM) \cite{Belkin2006,Melacci2011} is a special case
of GKM(s) when using Hinge loss and the smoothness function $l_{2}\left(.\right)$.
We then develop a new algorithm based on the SGD framework \cite{Lacoste2012}
to directly solve the optimization problem of GKM in its primal form
with the ideal convergence rate $\textrm{O}\left(\frac{1}{T}\right)$.
At each iteration, a labeled instance and an edge in the spectral
graph are randomly sampled. As the result, the computational cost
at each iteration is very economic, hence making the proposed method
efficient to deal with large-scale datasets while maintaining comparable
predictive performance.

To summarize, our contributions in this paper are as follows:\vspace{-1mm}

\begin{itemize}
\item We provide a novel view of jointly learning the kernel method with
a spectral graph for semi-supervised learning. Our proposed GKM enables
the combination of a wide spectrum of convex loss functions and smoothness
functions. 
\item We use the stochastic gradient descent (SGD) to solve directly GKM
in its primal form. Hence, GKM has all advantageous properties of
SGD-based methods including fast computation, memory efficiency, and
online setting. To the best of our knowledge, the proposed GKM is
\emph{the first kernelized semi-supervised learning method} that can
deal with the \emph{online learning context} for large-scale application.
\item We provide a theoretical analysis to show that GKM has the ideal convergence
rate $\text{O}\left(\frac{1}{T}\right)$ if the function is smooth
and the loss function satisfies a proper condition. We then verify
that this necessary condition holds for a wide class of loss functions
including Hinge, smooth Hinge, and Logistic for classification task
and L1, $\varepsilon$-insensitive for regression task (cf. Section
\ref{sec:lossfunc}). 
\item The experimental results further confirm the ideal convergence rate
$\text{O}\left(\frac{1}{T}\right)$ of GKM empirically and show that
GKM is readily scalable for large-scale datasets. In particular, it
offers a comparable classification accuracy whilst achieving a significant
computational speed-up in comparison with the state-of-the-art baselines.
\end{itemize}

\vspace{-3mm}

\section{Related Work\label{sec:bg}}

We review the works in semi-supervised learning paradigm that are
closely related to ours. Graph-based semi-supervised learning is an
active research topic under semi-supervised learning paradigm. At
its crux, graph-based semi-supervised methods define a graph where
the vertices are labeled and unlabeled data of the training set and
edges (may be weighted) reflect the similarity of data. Most of graph-based
methods can be interpreted as estimating the prediction function $f$
such that: it should predict the labeled data as accurate as possible;
and it should be smooth on the graph. 

In \citep{blum2001,blum2004}, semi-supervised learning problem is
viewed as graph mincut problem. In the binary case, positive labels
act as \textit{sources} and negative labels act as \textit{sinks}.
The objective is to find a minimum set of edges whose removal blocks
all flow from the sources to the sinks. Another way to infer the labels
of unlabeled data is to compute the marginal probability of the discrete
Markov random field. In \citep{Zhu02towardssemi-supervised}, Markov
Chain Monte Carlo sampling technique is used to approximate this marginal
probability. The work of \citep{getz2006} proposes to compute the
marginal probabilities of the discrete Markov random field at any
temperature with the Multi-canonical Monte Carlo method, which seems
to be able to overcome the energy trap faced by the standard Metropolis
or Swendsen - Wang method. The harmonic functions used in \citep{zhu03}
is regarded as a continuous relaxation of the discrete Markov random
field. It does relaxation on the value of the prediction function
and makes use of the quadratic loss with infinite weight so that the
labeled data are clamped. The works of \citep{kondor2002diffusionkernel,Chapelle2003,SmolaKondor2003}
utilize the Laplacian matrix induced from the spectral graph to form
kernel functions which can capture the features of the ambient space.

Yet another successful approach in semi-supervised learning paradigm
is the kernel-based approach. The kernel-based semi-supervised methods
are primarily driven by the idea to solve a standard SVM problem while
treating the unknown labels as optimization variables \citep{Chapelle2008}.
This leads to a non-convex optimization problem with a combinatorial
explosion of label assignments. Many methods have been proposed to
solve this optimization problem, for example local combinatorial search
\citep{Joachims99}, gradient descent \citep{Chapelle2005}, continuation
techniques \citep{Chapelle2006}, convex-concave procedures \citep{Collobert06},
deterministic annealing \citep{Sindhwani2006,le2013,nguyen2014},
and semi-definite programming \citep{Bie06}. However, the requirement
of retraining the whole dataset over and over preludes the applications
of these kernel-based semi-supervised methods to the large-scale and
streaming real-world datasets. 

Some recent works on semi-supervised learning have primarily concentrated
on the improvements of its safeness and classification accuracy. \citet{Li2015}
assumes that the low-density separators can be diverse and an incorrect
selection may result in a reduced performance and then proposes S4VM
to use multiple low-density separators to approximate the ground-truth
decision boundary. S4VM is shown to be safe and to achieve the maximal
performance improvement under the low-density assumption of S3VM \citep{Joachims99}.
\citet{WangCXF15} extends \citep{Belkin2006,Wu20101641} to propose
semi-supervised discrimination-aware manifold regularization framework
which considers the discrimination of all available instances in learning
of manifold regularization. \citet{TanZDZ14} proposes using the
$p$-norm as a regularization quantity in manifold regularization
framework to perform the dimensionality reduction task in the context
of semi-supervised learning. 

The closest work to ours is the manifold regularization framework
\citep{Belkin2006} and its extensions \citep{Sindhwani05linearmanifold,TsangK06,Melacci2011}.
However, the original work of manifold regularization \citep{Belkin2006}
requires to invert a matrix of size $n$ by $n$ which costs cubically
and hence is not scalable. Addressing this issue, \citet{TsangK06}
scales up the manifold regularization framework by adding in an $\varepsilon$-insensitive
loss into the energy function, i.e., replacing $\sum w_{ij}\left(f\left(x_{i}\right)-f\left(x_{j}\right)\right)^{2}$
by $\sum w_{ij}\left(\left|f\left(x_{i}\right)-f\left(x_{j}\right)\right|_{\varepsilon}\right)^{2}$,
where $\left|z\right|_{\varepsilon}=\max\left\{ \left|z\right|-\varepsilon,0\right\} $.
The intuition is that most pairwise differences $\left|f\left(x_{i}\right)-f\left(x_{j}\right)\right|$
are very small. By ignoring the differences smaller than $\varepsilon$,
the solution becomes sparser. LapSVM (in primal) \citep{Melacci2011}
employs the preconditioned conjugate gradient to solve the optimization
problem of manifold regularization in the primal form. This allows
the computational complexity to be scaled up from $\textrm{O}\left(n^{3}\right)$
to $\textrm{O}\left(n^{2}\right)$. However, the optimization problem
in \citep{Melacci2011} is indeed solved in the first dual layer rather
than in the primal form. Furthermore, we empirically show that LapSVM
in primal is expensive in terms of memory complexity (cf. Table \ref{tab:balanced80}).

Finally, the preliminary results of this work has been published in
\citep{Le_2016Budgeted} where it presents a special case of this
work in which the combination of Hinge loss and the smoothness function
$l_{1}\left(\cdot\right)$ is employed. In addition, our preliminary
work \citep{Le_2016Budgeted} guarantees only the convergence rate
$\text{O}\left(\frac{\log\,T}{T}\right)$. In this work, we have substantially
expanded \citep{Le_2016Budgeted} and developed more powerful theory
that guarantees the ideal convergence rate. We then further developed
theory and analysis in order to enable the employment of a wide spectrum
of loss and smoothness functions. Finally, we expanded significantly
on the new technical contents, explanations as well as more extensive
empirical studies.

\vspace{-3mm}

\section{Graph Setting for Semi-supervised Learning }

\begin{figure}[t]
\begin{centering}
\includegraphics[width=0.7\columnwidth]{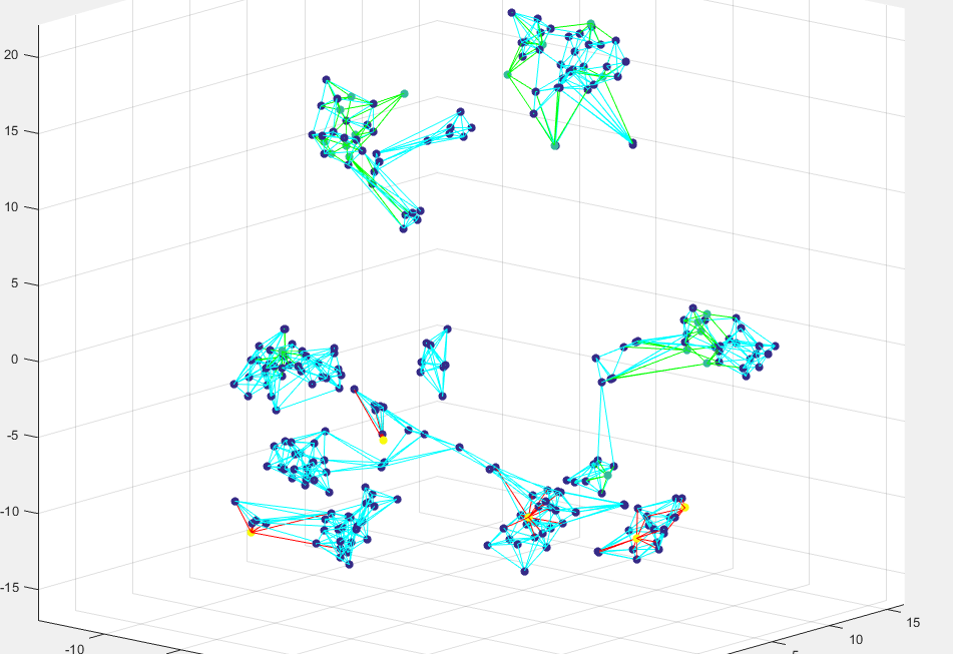}
\par\end{centering}
\caption{Visualization of a spectral graph on 3D dataset using $k$-NN with
$k=5$.\label{fig:sg}}
\end{figure}
Our \emph{spectral graph} is defined as a pair $\mathcal{G}=\left(\mathcal{V},\mathcal{E}\right)$
comprising a set $\mathcal{V}$ of vertices or nodes or points together
with a set $\mathcal{E}$ of edges or arcs or lines, which are 2-element
subsets of $V$ (i.e. an edge is associated with two vertices, and
that association takes the form of the unordered pair comprising those
two vertices). In the context of semi-supervised learning, we are
given a training set $X=X_{l}\cup X_{u}$ where $X_{l}=\left\{ \left(x_{i},y_{i}\right)\right\} _{i=1}^{l}$
is labeled data and $X_{u}=\left\{ x_{i}\right\} _{i=l+1}^{l+u}$
is unlabeled data. We construct the vertex set $\mathcal{V}$ includes
all labeled and unlabeled instances (i.e., $\mathcal{V}=\left\{ x_{i}\right\} _{i=1}^{l+u}$).
An edge $e_{ij}=\overline{x_{i}x_{j}}\in\mathcal{E}$ between two
vertices $x_{i},\,x_{j}$ represents the similarity of the two instances.
Let $\mu_{ij}$ be the weight associated with edge $e_{ij}$. The
underlying principle is to enforce that if $\mu_{ij}$ is large, then
$y_{i}$ and $y_{j}$ are expected to receive the same label. The
set of edges $\mathcal{G}$ and its weighs can be built using the
following ways \citep{Zhu2009,Le_2016Budgeted}:
\begin{itemize}
\item \textit{Fully connected graph}: every pair of vertices $x_{i}$, $x_{j}$
is connected by an edge. The edge weight decreases when the distance
$||x_{i}-x_{j}||$ increases. The Gaussian kernel function widely
used is given by $\mu_{ij}=\text{exp}\left(-\Vert x_{i}-x_{j}\Vert^{2}/\left(2\sigma_{s}^{2}\right)\right)$
where $\sigma_{s}$ controls how quickly the weight decreases.
\item $k$\textit{-NN}: each vertex $x_{i}$ determines its $k$ nearest
neighbors ($k$-NN) and makes an edge with each of its $k$-NN. The
Gaussian kernel weight function can be used for the edge weight. Empirically,
$k$-NN graphs with small $k$ tend to perform well.
\item $\varepsilon$\textit{-NN}: we connect $x_{i}$ and $x_{j}$ if $\norm{x_{i}-x_{j}}\leq\varepsilon$.
Again the Gaussian kernel weight function can be used to weight the
connected edges. In practice, $\varepsilon$-NN graphs are easier
to construct than $k$-NN graphs.
\end{itemize}
When constructing a graph, we avoid connecting the edge of two labeled
instances since we do not need to propagate the label between them.
Figure \ref{fig:sg} illustrates an example of  graph constructed
on 3D dataset using $k$-NN with $k=5$.

After constructing a graph, we formulate a semi-supervised learning
problem as assigning labels to the unlabeled vertices via \emph{label
propagation}. We propagate the labels from the labeled vertices to
the unlabeled ones by encouraging $y_{i}$ to have the same label
as $y_{j}$ if the weight $\mu_{ij}$ is large and vice versa. To
do so, we learn a mapping function $f:\,\mathfrak{\mathcal{X}}\rightarrow\mathcal{Y}$
where $\mathcal{X}$ and $\mathcal{Y}$ are the domains of data and
label such that
\begin{itemize}
\item $f\left(x_{i}\right)$ is as closest to its label $y_{i}$ as possible
for all labeled instances $x_{i}\,\left(1\leq i\leq l\right)$.
\item $f$ should be smooth on the graph $\mathcal{G}$, i.e., if $x_{i}$
is very close to $x_{j}$ (i.e., $\mu_{ij}$ is large), the discrepancy
between $f_{i}$ and $f_{j}$ (i.e., $|f_{i}-f_{j}|$) is small.
\end{itemize}
Therefore, we obtain the optimization problem \citep{Zhu2009,Le_2016Budgeted}
as 
\begin{equation}
\underset{f}{\min}\,\left(\infty\times\sum_{i=1}^{l}\left(f\left(x_{i}\right)-y_{i}\right)^{2}+\sum_{\left(i,j\right)\in\mathcal{E}}\mu_{ij}\left(f\left(x_{i}\right)-f\left(x_{j}\right)\right)^{2}\right)\label{eq:gen_op}
\end{equation}

\noindent where by convention we define $\infty\times0=0$. The optimization
problem in Eq. (\ref{eq:gen_op}) get the minimum when the first term
is $0$ and the second term is as smallest as possible. We rewrite
the optimization problem in Eq. (\ref{eq:gen_op}) into a constrained
optimization problem to link it to SVM-based optimization problem
for semi-supervised learning:
\begin{align}
\underset{f}{\min}\,\left(\sum_{\left(i,j\right)\in\mathcal{E}}\mu_{ij}\left(f\left(x_{i}\right)-f\left(x_{j}\right)\right)^{2}\right) & \thinspace\thinspace\thinspace\thinspace\thinspace\thinspace\thinspace\thinspace\thinspace\thinspace\thinspace\thinspace\thinspace\thinspace\thinspace\thinspace\thinspace\thinspace\thinspace\text{s.t.}:\,\forall_{i=1}^{l}:f\left(x_{i}\right)=y_{i}\thinspace\thinspace\thinspace\thinspace\thinspace\thinspace\thinspace\thinspace\thinspace\thinspace\thinspace\thinspace\thinspace\label{eq:gen_op2}
\end{align}
To extend the representation ability of the prediction function $f$,
we relax the discrete function $f$ to be real-valued. The drawback
of the relaxation is that in the solution, $f(x)$ is now real-valued
and hence does not directly correspond to a label. This can however
be addressed by thresholding $f(x)$ at zero to produce discrete label
predictions, i.e., if $f(x)\geq0$, predict $y=1$, and if $f(x)<0$,
predict $y=-1$.
\vspace{-3mm}

\section{Graph-based Semi-supervised Kernel Machine}

In this section, we present our proposed \emph{Graph-based Semi-supervised
Kernel Machine} (GKM). We begin with formulating the optimization
problem for GKM, followed by the derivation of an SGD-based solution.
Finally, we present the convergence analysis for our proposed method.
The convergence analysis shows that our proposed method gains the
ideal convergence rate $\text{O}\left(\frac{1}{T}\right)$ for combinations
of the typical loss functions (cf. Section \ref{sec:lossfunc}) and
the smoothness functions with $p\geq1$.

\subsection{GKM Optimization Problem}

Let $\Phi:\mathcal{X}\rightarrow\mathcal{H}$ be a transformation
from the input space $\mathcal{X}$ to a RHKS $\mathcal{H}$. To predict
label, we use the function $f\left(\boldsymbol{x}\right)=\transp{\bw}\Phi\left(x\right)-\rho=\sum_{i=1}^{l+u}\alpha_{i}K\left(x_{i},x\right)-\rho$,
where $\bw=\sum_{i=1}^{l+u}\alpha_{i}\Phi\left(x_{i}\right)$. Continuing
from the constrained optimization problem in Eq. (\ref{eq:gen_op2}),
we propose the following optimization problem over a graph \citep{Le_2016Budgeted}
\begin{gather}
\underset{\bw}{\textrm{min}}\,\left(\frac{1}{2}\left\Vert \bw\right\Vert ^{2}+\frac{C}{l}\sum_{i=1}^{l}\xi_{i}+\frac{C'}{|\mathcal{E}|}\sum_{\left(i,j\right)\in\mathcal{E}}\mu_{ij}\left(f\left(x_{i}\right)-f\left(x_{j}\right)\right)^{2}\right)\label{eq:gb_s3vm_op}\\
\text{s.t.}:\,\forall_{i=1}^{l}:y_{i}\left(\transp{\bw}\Phi\left(x_{i}\right)-\rho\right)\geq1-\xi_{i};\thinspace\thinspace\thinspace\thinspace\thinspace\thinspace\thinspace\thinspace\thinspace\thinspace\thinspace\thinspace\thinspace\textrm{and}\thinspace\thinspace\thinspace\thinspace\thinspace\thinspace\thinspace\thinspace\forall_{i=1}^{l}:\xi_{i}\geq0\nonumber 
\end{gather}

\noindent where $f\left(x_{i}\right)=\transp{\bw}\Phi\left(\boldsymbol{x_{i}}\right)-\rho$.
The optimization problem in Eq. (\ref{eq:gb_s3vm_op}) can be intuitively
understood as follows. We minimize $\frac{1}{2}\left\Vert \boldsymbol{\bw}\right\Vert ^{2}$
to maximize the margin to promote the generalization capacity, similar
to the intuition of SVM \citep{svm_original}. At the same time, we
minimize $\sum_{\left(i,j\right)\in\mathcal{E}}\mu_{ij}\left(f\left(x_{i}\right)-f\left(x_{j}\right)\right)^{2}$
to make the prediction function smoother on the spectral graph. We
rewrite the optimization problem in Eq. (\ref{eq:gb_s3vm_op}) in
the primal form as\footnote{We can eliminate the bias $\rho$ by simply adjusting the kernel.} 

\noindent\begin{equation}
\resizebox{.91\linewidth}{!}{\noindent\begin{minipage}{\linewidth}
\begin{equation*}
\underset{\bw}{\min}\left(\frac{\norm{\bw}^{2}}{2}+\frac{C}{l}\sum_{i=1}^{l}l\left(\bw;z_{i}\right)+\frac{C'}{|\mathcal{E}|}\sum_{\left(i,j\right)\in\mathcal{E}}\mu_{ij}l_{2}\left(\transp{\bw}\boldsymbol{\Phi}_{ij}\right)\right)\end{equation*} \end{minipage} }\label{eq:primal} 
\end{equation}

\noindent where $z_{i}=\left(x_{i},y_{i}\right)$, $l\left(\bw;x,y\right)=\max\left\{ 0,1-y\transp{\bw}\Phi\left(x\right)\right\} $,
$\boldsymbol{\Phi}_{ij}=\Phi\left(x_{i}\right)-\Phi\left(x_{j}\right)$,
$l_{p}\left(t\right)=\vert t\vert^{p}$ with $t\in\mathbb{R}$, and
$p\geq1$. In the optimization problem in Eq. (\ref{eq:primal}),
the minimization of $\sum_{i=1}^{l}l\left(\bw;x_{i},y_{i}\right)$
encourages the fitness of GKM on the labeled portion while the minimization
of $\sum_{\left(i,j\right)\in\mathcal{E}}\mu_{ij}l_{2}\left(\transp{\bw}\boldsymbol{\Phi}_{ij}\right)$
guarantees the smoothness of GKM on the spectral graph. Naturally,
we can extend the optimization of GKM by replacing the Hinge loss
by any loss function and $l_{2}\left(.\right)$ by $l_{p}\left(.\right)$
with $p\geq1$. 

We note that Laplacian Support Vector Machine (LapSVM) \citep{Belkin2006,Melacci2011}
is a special case of GKM using the Hinge loss with the smoothness
function $l_{2}\left(.\right)$. 

\subsection{Stochastic Gradient Descent Algorithm for GKM}

We employ the SGD framework \citep{Lacoste2012} to solve the optimization
problem in Eq. (\ref{eq:primal}) in the primal form. Let us denote
the objective function as
\[
\mathcal{J}\left(\bw\right)\triangleq\frac{\norm{\bw}^{2}}{2}+\frac{C}{l}\sum_{i=1}^{l}l\left(\bw;x_{i},y_{i}\right)+\frac{C'}{|\mathcal{E}|}\sum_{\left(i,j\right)\in\mathcal{E}}\mu_{ij}l_{p}\left(\transp{\bw}\boldsymbol{\Phi}_{ij}\right)
\]
At the iteration $t$, we do the following:
\begin{itemize}
\item Uniformly sample a labeled instance $x_{i_{t}}$ ($1\leq i_{t}\leq l$)
from the labeled portion $X_{l}$ and an edge $\left(u_{t},v_{t}\right)$
from the set of edges $\mathcal{E}$.
\item Define the instantaneous objective function 
\[
\mathcal{J}_{t}\left(\bw\right)=\frac{\norm{\bw}^{2}}{2}+Cl\left(\bw;x_{i_{t}},y_{i_{t}}\right)+C'\mu_{u_{t}v_{t}}l_{p}\left(\transp{\bw}\boldsymbol{\Phi}_{u_{t}v_{t}}\right)
\]
\item Define the stochastic gradient $g_{t}=\nabla_{\bw}\mathcal{J}_{t}\left(\bw_{t}\right)$
\begin{align*}
g_{t} & =\bw_{t}+C\nabla_{\bw}l\left(\bw_{t};x_{i_{t}},y_{i_{t}}\right)+C'\mu_{u_{t}v_{t}}\nabla_{\bw}l_{p}\left(\transp{\bw_{t}}\boldsymbol{\Phi}_{u_{t}v_{t}}\right)
\end{align*}
where $\nabla_{\bw}l\left(\bw;x,y\right)$ specifies the derivative
or sub-gradient w.r.t. $\bw$.
\item Update $\bw_{t+1}$ with the learning rate $\eta_{t}=\frac{2}{t+1}$,
$\bw_{t+1}=\bw_{t}-\eta_{t}g_{t}$
\begin{align*}
\bw_{t+1} & =\frac{t-1}{t+1}\bw_{t}-\frac{2C}{t+1}\nabla_{\bw}l\left(\bw_{t};x_{i_{t}},y_{i_{t}}\right)-\frac{2C'\mu_{u_{t}v_{t}}}{t+1}\nabla_{\bw}l_{p}\left(\transp{\bw_{t}}\boldsymbol{\Phi}_{u_{t}v_{t}}\right)
\end{align*}
\item Update $\overline{\bw}_{t+1}=\frac{t-1}{t+1}\overline{\bw}_{t}+\frac{2}{t+1}\bw_{t+1}$
\end{itemize}
We note that the derivative $\nabla_{\bw}l_{p}\left(\transp{\bw}\boldsymbol{\Phi}\right)$
w.r.t. $\bw$ can be computed as
\[
\nabla_{\bw}l_{p}\left(\transp{\bw}\boldsymbol{\Phi}\right)=p\text{sign}\left(\transp{\bw}\boldsymbol{\Phi}\right)\vert\transp{\bw}\boldsymbol{\Phi}\vert^{p-1}\boldsymbol{\Phi}
\]

\begin{algorithm}[H]
\caption{Algorithm for GKM.\label{alg:gUS3VM}}

\textbf{Input :} $C$, $C'$, $p$ , $K\left(.,.\right)$
\begin{algor}[1]
\item [{{*}}] $\bw_{1}=\bzero$, $\overline{\bw}_{1}=\bzero$
\item [{for}] $t=1$ \textbf{to} $T$
\item [{{*}}] Uniformly sample $i_{t}$ from \{$1,2,...,l$\} and $\left(u_{t},v_{t}\right)$
from the set of edges $\mathcal{E}$
\item [{{*}}] Update $\bw_{t+1}=\frac{t-1}{t+1}\bw_{t}-\frac{2C}{t+1}\nabla_{\bw}l\left(\bw_{t};x_{i_{t}},y_{i_{t}}\right)-\frac{2C'\mu_{u_{t}v_{t}}}{t+1}\nabla_{\bw}l_{p}\left(\transp{\bw_{t}}\boldsymbol{\Phi}_{u_{t}v_{t}}\right)$
\item [{{*}}] Update $\overline{\bw}_{t+1}=\frac{t-1}{t+1}\overline{\bw}_{t}+\frac{2}{t+1}\bw_{t+1}$
\item [{endfor}]~
\end{algor}
\textbf{Output :} $\overline{\bw}_{T+1}$
\end{algorithm}

The pseudocode of GKM is presented in Algorithm \ref{alg:gUS3VM}.
We note that we store $\bw_{t}$ and $\overline{\bw}_{t}$ as $\bw_{t}=\sum_{i}\alpha_{i}\Phi\left(x_{i}\right)$
and $\overline{\bw}_{t}=\sum_{i}\beta_{i}\Phi\left(x_{i}\right)$.
In line 5 of Algorithm \ref{alg:gUS3VM}, the update of $\bw_{t+1}$
involves the coefficients of $\Phi\left(x_{i_{t}}\right)$, $\Phi\left(x_{u_{t}}\right)$,
and $\Phi\left(x_{v_{t}}\right)$. In line 4 of Algorithm \ref{alg:gUS3VM},
we need to sample the edge $\left(u_{t},v_{t}\right)$ from the set
of edges $\mathcal{E}$ and compute the edge weight $\mu_{u_{t}v_{t}}$.
It is noteworthy that in GKM we use the fully connected spectral graph
to maximize the freedom of label propagation and avoid the additional
computation incurred in other kind of spectral graph (e.g., $k$-NN
or $\varepsilon$-NN spectral graph). In addition, the edge weight
$\mu_{u_{t}v_{t}}$ can be computed on the fly when necessary.

\subsection{Convergence Analysis \label{subsec:Convergence-Analysis}}

After presenting the SGD algorithm for our proposed GKM, we now present
the convergence analysis. In particular, assuming that the loss function
satisfies the condition: $\norm{\nabla_{\bw}l\left(\bw;x,y\right)}\leq A,\,\forall x,y$,
we prove that our GKM achieves the ideal convergence rate $\text{O}\left(\frac{1}{T}\right)$
with $1\leq p<2$ and with $p\geq2$ under some condition of the parameters
(cf. Theorem \ref{thm:regret}) while we note that the previous work
of \citep{Le_2016Budgeted} achieves a convergence rate of $\text{O}\left(\frac{\log T}{T}\right)$.
We present the theoretical results and the rigorous proofs are given
in Appendix \ref{sec:Convex-Analysis}. Without loss of generality,
we assume that the feature map $\Phi\left(x\right)$ is bounded in
the feature space, i.e., $\norm{\Phi\left(x\right)}=K\left(x,x\right)^{1/2}\leq R,\,\forall x$.
We denote the optimal solution by $\bw^{*}$, i.e., $\bw^{*}=\text{argmin}_{\vectorize w}\,J\left(\bw\right)$.

The following lemma shows the formula for $\overline{\bw}_{t}$ from
its recursive formula.

\begin{lemma}\label{lem:w_t}We have the following statement $\overline{\bw}_{t}=\frac{2}{t\left(t-1\right)}\sum_{i=1}^{t-1}i\bw_{i}$.

\end{lemma}Lemma \ref{lem:func} further offers the foundation to
establish an upper bound for $\norm{\bw_{t}}$.

\begin{lemma}\label{lem:func}Let us consider the function $f\left(x;a,b,p\right)=ax^{p-1}-x+b$
where $x\geq0$ and $p\geq1$, $a,\,b>0$. The following statements
are guaranteed

i) If $p<2$ then $f\left(M;a,b,p\right)\leq0$ where $M=\max\left(1,\left(a+b\right)^{\frac{1}{2-p}}\right)$.

ii) If $p=2$ and $a<1$ then $f\left(M;a,b,p\right)\leq0$ where
$M=\frac{b}{1-a}$.

iii) If $p>2$ and $ab^{p-2}\leq\frac{\left(p-2\right)^{p-2}}{\left(p-1\right)^{p-1}}$
then $f\left(M;a,b,p\right)\leq0$ where $M=\left(\frac{1}{\left(p-1\right)a}\right)^{\frac{1}{p-2}}$.

\end{lemma}Built on the previous lemma, Lemma \ref{lem:wt_norm}
establishes an upper bound on $\norm{\bw_{t}}$ which is used to define
the bound in Lemma \ref{lem:gt}.

\begin{lemma}\label{lem:wt_norm}We have the following statement
$\norm{\bw_{t}}\leq M,\,\forall t$ where $M$ is defined as 
\[
M=\begin{cases}
\max\left(1,\left(a+b\right)^{\frac{1}{2-p}}\right) & \text{if}\,p<2\\
\frac{b}{1-a} & \text{if}\,p=2,a<1\\
\left(\frac{1}{\left(p-1\right)a}\right)^{\frac{1}{p-2}} & \text{if}\,p>2,ab^{p-2}\leq\frac{\left(p-2\right)^{p-2}}{\left(p-1\right)^{p-1}}
\end{cases}
\]
with $a=C'\left(2R\right)^{p}p$ and $b=CA$.\end{lemma}Lemma \ref{lem:gt}
establishes an upper bound on $\norm{g_{t}}$ for our subsequent theorems.

\begin{lemma}\label{lem:gt}We have $\norm{g_{t}}\leq G,\,\forall t$
where $G=M+CA+C'\left(2R\right)^{p}pM^{p-1}$.\end{lemma}We now turn
to establish the ideal convergence rate $\text{O}\left(\frac{1}{T}\right)$
for our proposed GKM.

\begin{theorem}\label{thm:regret}Considering the running of Algorithm
\ref{alg:gUS3VM}, the following statements hold

i) $\mathbb{E}\left[\mathcal{J}\left(\overline{\bw}_{T+1}\right)\right]-\mathcal{J}\left(\bw^{*}\right)\leq\frac{2G^{2}}{T},\,\forall T$

ii) $\mathbb{E}\left[\norm{\overline{\bw}_{T+1}-\bw^{*}}^{2}\right]\leq\frac{4G^{2}}{T},\,\forall T$ 

\noindent if $1\leq p<2$ or $p=2,\,a<1$ or $p>2,\,ab^{p-2}\leq\frac{\left(p-2\right)^{p-2}}{\left(p-1\right)^{p-1}}$
where $a=C'\left(2R\right)^{p}p$ and $b=CA$.

\end{theorem}Theorem \ref{thm:regret} states the regret in the form
of expectation. We go further to prove that for all $T\geq T_{0}=\left\lceil \frac{2G^{2}}{\varepsilon\delta}\right\rceil $,
with a high confidence level, $\mathcal{J}\left(\overline{\bw}_{T+1}\right)$
approximates the optimal value $\mathcal{J}\left(\bw^{*}\right)$
within an $\varepsilon$-precision in Theorem \ref{thm:high_prob_J}.

\begin{theorem}\label{thm:high_prob_J}With the probability $1-\delta$,
$\forall T\geq T_{0}=\left\lceil \frac{2G^{2}}{\varepsilon\delta}\right\rceil $,
$\mathcal{J}\left(\overline{\bw}_{T+1}\right)$ approximates the optimal
value $\mathcal{J}\left(\bw^{*}\right)$ within $\varepsilon$-precision,
i.e., $\mathcal{J}\left(\overline{\bw}_{T+1}\right)\leq\mathcal{J}\left(\bw^{*}\right)+\varepsilon$
if $1\leq p<2$ or $p=2,\,a<1$ or $p>2,\,ab^{p-2}\leq\frac{\left(p-2\right)^{p-2}}{\left(p-1\right)^{p-1}}$
where $a=C'\left(2R\right)^{p}p$ and $b=CA$.

\end{theorem}
\vspace{-3mm}

\section{Suitability of Loss Function and Kernel Function\label{sec:lossfunc}}

In this section, we present the suitability of five loss and kernel
functions that can be used for GKM including hinge, smooth hinge,
logistic, L1, and $\epsilon$-insensitive. We verify that most of
the well-known loss functions satisfy the necessary condition: $\norm{\nabla_{\bw}l\left(\bw;x,y\right)}\leq A$
for an appropriate positive number $A$. By allowing multiple choices
of loss functions, we enable broader applicability of the proposed
GKM for real-world applications while the work of \citep{Le_2016Budgeted}
is restricted to a Hinge loss function.
\begin{itemize}
\item \textit{Hinge loss}: $l\left(\bw;x,y\right)=\max\left\{ 0,1-y\transp{\bw}\Phi\left(x\right)\right\} $,
\begin{align*}
\nabla_{\bw}l\left(\bw;x,y\right) & =-\mathbb{I}_{\left\{ y\transp{\bw}\Phi\left(x\right)\leq1\right\} }y\Phi\left(x\right)
\end{align*}
Therefore, by choosing $A=R$ we have $\norm{\nabla_{\bw}l\left(\bw;x,y\right)}\leq\norm{\Phi\left(x\right)}\leq R=A$.
\item \textit{Smooth Hinge loss}\textbf{ }\citep{Shalev-Shwartz013}:\textbf{
}
\begin{align*}
l\left(\bw;x,y\right) & =\begin{cases}
0 & \text{if}\,\,yo>1\\
1-y\bw^{\top}\Phi\left(x\right)-\frac{\tau}{2} & \text{if}\,\,yo<1-\tau\\
\frac{1}{2\tau}\left(1-yo\right)^{2} & \text{otherwise}
\end{cases}\\
\nabla_{\bw}l\left(\bw;x,y\right) & =-\mathbb{I}_{\left\{ yo<1-\tau\right\} }y\Phi\left(x\right)+\tau^{-1}\mathbb{I}_{1-\tau\leq yo\leq1}\left(yo-1\right)y\Phi\left(x\right)
\end{align*}
where $o=\bw^{\top}\Phi\left(x\right)$. Therefore, by choosing $A=R$
we have
\begin{align*}
\norm{\nabla_{\bw}l\left(\bw;x,y\right)} & =R\left|\tau^{-1}\mathbb{I}_{1-\tau\leq yo\leq1}\left(yo-1\right)\right|+R\mathbb{I}_{\left\{ yo<1-\tau\right\} }\\
 & \leq\mathbb{I}_{\left\{ yo<1-\tau\right\} }R+\tau^{-1}\tau\mathbb{I}_{1-\tau\leq yo\leq1}R\leq R=A
\end{align*}
\item \textit{Logistic loss}: $l\left(\bw;x,y\right)=\log\left(1+\exp\left(-y\transp{\bw}\Phi\left(x\right)\right)\right)$,
\begin{align*}
\nabla_{\bw}l\left(\bw;x,y\right) & =\frac{-y\exp\left(-y\transp{\bw}\Phi\left(x\right)\right)\Phi\left(x\right)}{\exp\left(-y\transp{\bw}\Phi\left(x\right)\right)+1}
\end{align*}
By choosing $A=R$ we have $\norm{\nabla_{\bw}l\left(\bw;x,y\right)}<\norm{\Phi\left(x\right)}\leq R=A$.
\item \textit{L1 loss}: $l\left(\bw;x,y\right)=\vert y-\transp{\bw}\Phi\left(x\right)\vert$,
$\nabla_{\bw}l\left(\bw;x,y\right)=\text{sign}\left(\transp{\bw}\Phi\left(x\right)-y\right)\Phi\left(x\right)$.
By choosing $A=R$ we have $\norm{\nabla_{\bw}l\left(\bw;x,y\right)}\leq\norm{\Phi\left(x\right)}\leq R=A$.
\item $\varepsilon$-\textit{insensitive loss}: let denote $o=\transp{\bw}\Phi\left(x\right)\vert$,
$l\left(\bw;x,y\right)=\max\left\{ 0,\vert y-\transp{\bw}\Phi\left(x\right)\vert-\varepsilon\right\} $,
\begin{align*}
\nabla_{\bw}l\left(\bw;x,y\right) & =\mathbb{I}_{\left\{ \vert y-o>\varepsilon\right\} }\text{sign}\left(o-y\right)\Phi\left(x\right)
\end{align*}
By choosing $A=R$ we have $\norm{\nabla_{\bw}l\left(\bw;x,y\right)}\leq\norm{\Phi\left(x\right)}\leq R=A$.
\end{itemize}
Here, $\mathbb{I}_{S}$ is the indicator function which is equal to
$1$ if the statement $S$ is true and $0$ if otherwise. It can be
observed that the positive constant $A$ coincides the radius $R$
(i.e., $A=R$) for the aforementioned loss functions. To allow the
ability to flexibly control the minimal sphere that encloses all $\Phi\left(x\right)$(s),
we propose to use the squared exponential (SE) kernel function $k\left(x,x'\right)=\sigma_{f}^{2}\text{exp}\left(-\norm{x-x'}^{2}/\left(2\sigma_{l}^{2}\right)\right)$
where $\sigma_{l}$ is the length-scale and $\sigma_{f}$ is the output
variance parameter. Using SE kernel, we have the following equality
$\norm{\Phi\left(x\right)}=K\left(x,x\right)^{1/2}=\sigma_{f}\leq R$.
Recall that if $p=2$ or $p>2$, Algorithm \ref{alg:gUS3VM} converges
to the optimal solution with the ideal convergence rate $\text{O}\left(\frac{1}{T}\right)$
under specific conditions. In particular, with $p=2$ the corresponding
condition is $a<1$ and with $p>2$ the corresponding condition is
$ab^{p-2}\leq\frac{\left(p-2\right)^{p-2}}{\left(p-1\right)^{p-1}}$
where $a=C'\left(2R\right)^{p}p$ and $b=CA$. Using the SE kernel,
we can adjust the output variance parameter $\sigma_{f}$ to make
the convergent condition valid. More specifically, we consider two
cases:
\begin{itemize}
\item $p=2$: we have $a<1\goto C'\left(2R\right)^{p}p<1\goto R<0.5\left(pC'\right)^{-1/p}$.
We can simply choose $\sigma_{f}$ with a very small number $\rho>0$
\begin{align*}
\sigma_{f} & =R=0.5\left(pC'\right)^{-1/p}-\rho=0.5\left(2C'\right)^{-1/2}-\rho
\end{align*}
\item $p>2$: we have $ab^{p-2}\leq\left(p-2\right)^{p-2}\left(p-1\right)^{1-p}$
and then
\begin{align*}
C'\left(2R\right)^{p}p\left(CA\right)^{p-2} & \leq\left(p-2\right)^{p-2}\left(p-1\right)^{1-p}\\
R & \leq\left(\frac{\left(p-2\right)^{p-2}}{2^{p}C^{p-2}C'p\left(p-1\right)^{p-1}}\right)^{\frac{1}{2p-2}}
\end{align*}
We can simply choose $\sigma_{f}=R=\left(\frac{\left(p-2\right)^{p-2}}{2^{p}C^{p-2}C'p\left(p-1\right)^{p-1}}\right)^{\frac{1}{2p-2}}$.
\end{itemize}
We can control the second trade-off parameter $C'$ to ensure the
ideal convergence rate $\text{O}\left(\frac{1}{T}\right)$ with $p\geq2$.
More specifically, we also consider two cases:
\begin{itemize}
\item $p=2$: we have $a<1\goto C'<\left(2R\right)^{-p}p^{-1}=0.125$.
\item $p>2$: we have $ab^{p-2}\leq\left(p-2\right)^{p-2}\left(p-1\right)^{1-p}$
and thus 
\begin{align}
C' & \leq\frac{\left(p-2\right)^{p-2}}{2^{p}C^{p-2}p\left(p-1\right)^{p-1}}\label{eq:pgt2_RBF}
\end{align}
\end{itemize}

\section{Experiments}

\vspace{-10pt}
\begin{wraptable}{r}{0.5\columnwidth}%
\vspace{-18pt}

\begin{centering}
\begin{tabular}{|c|c|c|}
\hline 
\textbf{Dataset} & \textbf{Size} & \textbf{Dimension}\tabularnewline
\hline 
COIL20 & 145 & 1,014\tabularnewline
\hline 
G50C & 551 & 50\tabularnewline
\hline 
USPST & 601 & 256\tabularnewline
\hline 
AUSTRALIAN  & 690 & 14\tabularnewline
\hline 
A1A & 1,605  & 123\tabularnewline
\hline 
SVMGUIDE3 & 1,243 & 21\tabularnewline
\hline 
SVMGUIDE1 & 3,089 & 4\tabularnewline
\hline 
MUSHROOMS & 8,124 & 112\tabularnewline
\hline 
W5A  & 9,888 & 300\tabularnewline
\hline 
W8A & 49,749  & 300\tabularnewline
\hline 
IJCNN1 & 49,990 & 22\tabularnewline
\hline 
COD-RNA & 59,535 & 8\tabularnewline
\hline 
COVTYPE & 581,012 & 54\tabularnewline
\hline 
\end{tabular}
\par\end{centering}
\begin{centering}
\par\end{centering}
\caption{Statistics for datasets used.\label{tab:detail}}

\vspace{-15pt}
\end{wraptable}%
We conduct the extensive experiments to investigate the influence
of the model parameters and other factors to the model behavior and
to compare our proposed GKM with the state-of-the-art baselines on
the benchmark datasets. In particular, we design three kinds of experiments
to analyze the influence of factors (e.g., the loss function, the
smoothness function, and the percentage of unlabeled portion) to the
model behavior. In the first experiment (cf. Section \ref{subsec:exp1}),
we empirically demonstrate the theoretical convergence rate $\text{O}\left(\frac{1}{T}\right)$
for all combinations of loss function (Hinge and Logistic) and smoothness
function ($p=1,2,3$) and also investigate how the number of iterations
affects the classification accuracy. In the second experiment (cf.
Section \ref{subsec:exp2}), we study the influence of the loss function
and the smoothness function to the predictive performance and the
training time when the percentage of unlabeled portion is either $80\%$
or $90\%$. In the third experiment (cf. Section \ref{subsec:exp3}),
we examine the proposed method under the semi-supervised setting where
the proportion of unlabeled data is varied from $50\%$ to $90\%$.
Finally, we compare our proposed GKM with the state-of-the-art baselines
on the benchmark datasets.

\paragraph{Datasets}

We use $13$ benchmark datasets\footnote{Most of the experimental datasets can be conveniently downloaded from
the URL https://www.csie.ntu.edu.tw/\textasciitilde{}cjlin/libsvmtools/datasets/.} (see Table \ref{tab:detail} for details) for experiments on semi-supervised
learning.

\paragraph{Baselines \label{subsec:Baselines}}

To investigate the efficiency and accuracy of our proposed method,
we make comparison with the following baselines which, to the best
of our knowledge, represent the state-of-the-art semi-supervised learning
methods:
\begin{itemize}
\item \emph{LapSVM in primal} \citep{Melacci2011}: Laplacian Support Vector
Machine in primal is a state-of-the-art approach in semi-supervised
classification based on manifold regularization framework. It can
reduce the computational complexity of the original LapSVM \citep{Belkin2006}
from $\text{O}\left(n^{3}\right)$ to $\text{O}\left(kn^{2}\right)$
using the preconditioned conjugate gradient and an early stopping
strategy.
\item \emph{CCCP} \citep{Collobert06}: It solves the non-convex optimization
problem encountered in the kernel semi-supervised approach using convex-concave
procedures.
\item \emph{Self-KNN} \citep{Zhu2009}: Self-training is one of the most
classical technique used in semi-supervised classification. Self-KNN
employs $k-$NN method as a core classifier for self-training.
\item \emph{SVM}: Support Vector Machine which is implemented using LIBSVM
solver \citep{libsvm} and trained with fully label setting. We use
fully labeled SVM as a milestone to judge how good the semi-supervised
methods are.
\end{itemize}
All compared methods are run on a Windows computer with the configuration
of CPU Xeon 3.47 GHz and 96GB RAM. All codes of baseline methods are
obtained from the corresponding authors.\vspace{-1mm}

\subsection{Model Analysis}

\subsubsection{Convergence Rate Analysis\label{subsec:exp1}}

\begin{figure}
\begin{centering}
\includegraphics[width=1\linewidth]{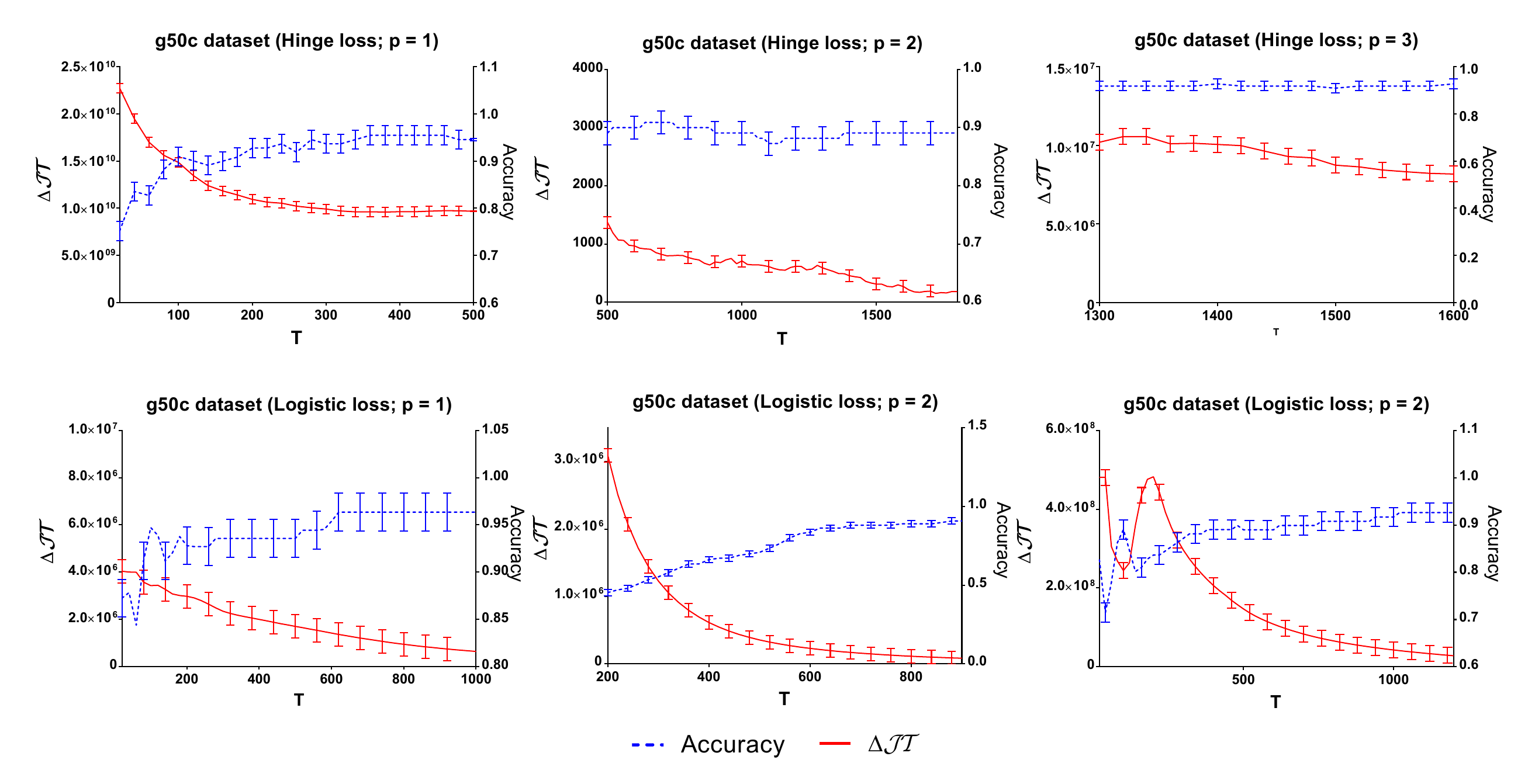}
\par\end{centering}
\caption{Convergence analysis on the dataset G50C using the accuracy and the
quantity $\Delta\mathcal{JT}=\left(\mathcal{J}\left(\overline{\protect\bw}_{T+1}\right)-\mathcal{J}\left(\protect\bw^{*}\right)\right)\times T$
. Hinge and Logistic losses are combined with $p=1,2,3$. When $T$
increases, the accuracy gradually improves and the quantity $\Delta\mathcal{JT}$
decreases to a constant.\label{fig:g50c_jt}}

\end{figure}

We empirically examine the convergence rate of GKM with various combinations
of loss functions (Hinge, Logistic) and smooth functions ($p=1,2,3$).
We select G50C dataset which we compute the quantity $\Delta\mathcal{JT}=\left(\mathcal{J}\left(\overline{\bw}_{T+1}\right)-\mathcal{J}\left(\bw^{*}\right)\right)\times T$
and measure the accuracy when the number of iterations $T$ is varied.
We repeat each experiment five times to record the necessary quantities
and their standard deviations. 

As observed from Figures \ref{fig:g50c_jt}, $\Delta\mathcal{JT}$
tends to decrease and when $T$ is sufficiently large, this quantity
is upper-bounded by a constant. Hence, we can empirically conclude
the convergence rate $\text{O}\left(\frac{1}{T}\right)$ of GKM. Empirical
result is consistent with the theoretical analysis developed in Section
\ref{subsec:Convergence-Analysis}. We use the RBF kernel and with
$p=2$ and $p=3$ the second trade-off parameter $C^{'}$ is selected
using Eqs. (\ref{eq:pgt2_RBF}) to theoretically guarantee the ideal
convergence rate $\text{O}\left(\frac{1}{T}\right)$ of our GKM.

\subsubsection{Loss Functions and Smoothness Functions Analysis\label{subsec:exp2}}

\begin{table}
\begin{centering}
\resizebox{1.0\textwidth}{!}{%
\begin{tabular}{c|cccccc|cccccc}
\hline 
\multirow{3}{*}{\textbf{Dataset}} & \multicolumn{6}{c|}{\textbf{Accuracy (\%)}} & \multicolumn{6}{c}{\textbf{Time (s)}}\tabularnewline
\cline{2-13} 
 & \multicolumn{3}{c}{\textbf{Hinge loss}} & \multicolumn{3}{c|}{\textbf{Logistic loss}} & \multicolumn{3}{c}{\textbf{Hinge loss}} & \multicolumn{3}{c}{\textbf{Logistic loss}}\tabularnewline
\cline{2-13} 
 & $p=1$ & $p=2$ & $p=3$ & $p=1$ & $p=2$ & $p=3$ & $p=1$ & $p=2$ & $p=3$ & $p=1$ & $p=2$ & $p=3$\tabularnewline
\hline 
A1A & 80.69 & 81.00 & 79.54 & 80.37 & 80.17 & 80.37 & 0.078 & 0.073 & 0.073 & 0.073 & 0.073 & 0.078\tabularnewline
AUSTRALIAN & 91.30 & 91.30 & 90.58 & 91.30 & 90.10 & 90.58 & 0.026 & 0.021 & 0.026 & 0.021 & 0.016 & 0.016\tabularnewline
COIL20 & 100 & 100 & 100 & 100 & 100 & 100 & 0.010 & 0.010 & 0.015 & 0.005 & 0.015 & 0.010\tabularnewline
G50C & 92.73 & 95.45 & 95.45 & 95.45 & 95.45 & 95.45 & 0.026 & 0.015 & 0.031 & 0.021 & 0.015 & 0.021\tabularnewline
SVM2 & 89.55 & 88.06 & 89.55 & 89.55 & 80.60 & 76.12 & 0.010 & 0.010 & 0.011 & 0.015 & 0.010 & 0.005\tabularnewline
SVM3 & 81.85 & 80.24 & 80.24 & 79.84 & 81.05 & 79.30 & 0.125 & 0.015 & 0.015 & 0.015 & 0.015 & 0.015\tabularnewline
USPST & 99.17 & 99.17 & 99.17 & 99.17 & 99.17 & 99.17 & 0.032 & 0.031 & 0.037 & 0.037 & 0.031 & 0.036\tabularnewline
COD-RNA & 87.25 & 87.43 & 86.32 & 87.05 & 87.04 & 87.26 & 1.432 & 0.995 & 0.891 & 1.011 & 1.094 & 1.573\tabularnewline
COVTYPE & 84.15 & 83.89 & 79.81 & 78.78 & 78.93 & 78.61 & 1.599 & 1.422 & 1.755 & 1.760 & 1.510 & 1.594\tabularnewline
IJCNN1 & 93.16 & 93.12  & 93.07 & 93.11 & 92.70 & 93.14 & 1.594 & 0.672 & 0.665 & 1.937 & 1.359 & 0.703\tabularnewline
W5A & 97.57 & 97.59 & 97.52 & 97.61 & 97.69 & 97.39 & 0.041 & 0.031 & 0.047 & 0.036 & 0.041 & 0.031\tabularnewline
W8A & 97.90 & 97.42 & 97.34 & 97.41 & 97.40 & 97.22 & 1.140 & 0.073 & 0.073 & 0.067 & 0.062 & 0.052\tabularnewline
MUSHROOM & 99.96 & 99.94 & 99.98 & 100 & 99.98 & 99.98 & 0.042 & 0.031 & 0.036 & 0.042 & 0.031 & 0.031\tabularnewline
\hline 
\end{tabular}}
\par\end{centering}
\caption{The classification performance comparison when the hidden labeled
data is $80\%$.\label{tab:our_alg_detail_acc_time_80}}
\end{table}

\begin{table}
\begin{centering}
\resizebox{1.0\textwidth}{!}{%
\begin{tabular}{c|cccccc|cccccc}
\hline 
\multirow{3}{*}{\textbf{Dataset}} & \multicolumn{6}{c|}{\textbf{Accuracy (\%)}} & \multicolumn{6}{c}{\textbf{Time (s)}}\tabularnewline
\cline{2-13} 
 & \multicolumn{3}{c}{\textbf{Hinge loss}} & \multicolumn{3}{c|}{\textbf{Logistic loss}} & \multicolumn{3}{c}{\textbf{Hinge loss}} & \multicolumn{3}{c}{\textbf{Logistic loss}}\tabularnewline
\cline{2-13} 
 & $p=1$ & $p=2$ & $p=3$ & $p=1$ & $p=2$ & $p=3$ & $p=1$ & $p=2$ & $p=3$ & $p=1$ & $p=2$ & $p=3$\tabularnewline
\hline 
A1A & 83.18 & 83.49 & 83.49 & 83.18 & 83.18 & 82.24 & 0.078 & 0.078 & 0.073 & 0.078 & 0.078 & 0.067\tabularnewline
AUSTRALIAN & 90.58 & 90.10 & 89.86 & 89.86 & 90.58 & 89.86 & 0.016 & 0.026 & 0.016 & 0.016 & 0.021 & 0.016\tabularnewline
COIL & 100 & 100 & 100 & 100 & 100 & 100 & 0.015 & 0.005 & 0.015 & 0.015 & 0.015 & 0.010\tabularnewline
G50C & 97.27 & 96.36 & 96.36 & 96.36 & 96.36 & 96.36 & 0.020 & 0.015 & 0.031 & 0.021 & 0.015 & 0.021\tabularnewline
SVM2 & 82.09 & 82.09 & 80.60 & 73.63 & 79.10 & 73.13 & 0.010 & 0.010 & 0.015 & 0.005 & 0.010 & 0.010\tabularnewline
SVM3 & 82.26 & 81.05 & 80.65 & 80.24 & 79.97 & 79.43 & 0.094 & 0.015 & 0.015 & 0.057 & 0.015 & 0.151\tabularnewline
USPST & 100 & 100 & 100 & 100 & 100 & 100 & 0.037 & 0.031 & 0.031 & 0.031 & 0.031 & 0.031\tabularnewline
COD-RNA & 87.37 & 87.51 & 87.07 & 87.38 & 87.23 & 86.18 & 0.974 & 1.115 & 0.896 & 1.078 & 1.000 & 0.865\tabularnewline
COVTYPE & 85.22 & 85.29 & 73.07 & 64.40 & 70.09 & 68.20 & 1.588 & 1.510 & 1.536 & 1.922 & 1.724 & 1.515\tabularnewline
IJCNN1 & 92.84 & 92.62 & 92.82 & 92.59 & 92.63 & 92.50 & 0.641 & 0.729 & 0.774 & 1.135 & 1.158 & 0.734\tabularnewline
W5A & 97.64 & 97.61 & 97.50 & 97.69 & 97.47 & 97.49 & 0.036 & 0.052 & 0.052 & 0.052 & 0.052 & 0.041\tabularnewline
W8A & 98.00 & 97.50 & 97.60 & 97.56 & 97.48 & 97.35 & 0.677 & 0.068 & 0.146 & 0.146 & 0.385 & 0.088\tabularnewline
MUSHROOM & 99.92 & 99.96 & 99.96 & 100 & 99.96 & 99.94 & 0.172 & 0.037 & 0.031 & 0.032 & 0.031 & 0.032\tabularnewline
\hline 
\end{tabular}}
\par\end{centering}
\caption{The classification performance comparison when the hidden labeled
data is $90\%$.\label{tab:our_alg_detail_acc_time_90}}
\end{table}

This experiment aims to investigate how the variation in loss function
and smoothness function affects the learning performance on real datasets.
We experiment on the real datasets given in Table \ref{tab:detail}
with different combinations of loss function (e.g., Hinge, Logistic)
and smoothness function (e.g., $p=1,2,3$). Each experiment is performed
five times and the average accuracy and training time corresponding
to $80\%$ and $90\%$ of unlabeled data are reported in Tables \ref{tab:our_alg_detail_acc_time_80}
and \ref{tab:our_alg_detail_acc_time_90} respectively. We observe
that the Hinge loss is slightly better than Logistic one and the combination
of Hinge loss and the smoothness $l_{1}\left(.\right)$ is the best
combination among others. It is noteworthy that in this simulation
study we use the RBF kernel and with $p=2$ and $p=3$ the second
trade-off parameter $C^{'}$ is selected using Eqs. (\ref{eq:pgt2_RBF})
to theoretically guarantee the ideal convergence rate $\text{O}\left(\frac{1}{T}\right)$
of our GKM.

\subsubsection{Unlabeled Data Proportion Analysis\label{subsec:exp3}}

\begin{table}[t]
\begin{centering}
\resizebox{1.0\textwidth}{!}{%
\begin{tabular}{c||c|c|c|c|c||c|c|c|c|c|}
\hline 
\multirow{2}{*}{\textbf{Dataset}} & \multicolumn{5}{c||}{\textbf{Accuracy (\%)}} & \multicolumn{5}{c|}{\textbf{F1 score (\%)}}\tabularnewline
\cline{2-11} 
 & \textbf{50\%} & \textbf{60\%} & \textbf{70\%} & \textbf{80\%} & \textbf{90\%} & \textbf{50\%} & \textbf{60\%} & \textbf{70\%} & \textbf{80\%} & \textbf{90\%}\tabularnewline
\hline 
G50C & 98.21 & 95.45 & 95.02 & 94.29 & 93.82 & 98.19 & 95.51 & 94.97 & 93.02 & 93.02\tabularnewline
\cline{2-11} 
\multirow{1}{*}{COIL20} & 100 & 100 & 100 & 100 & 100 & 100 & 100 & 100 & 100 & 100\tabularnewline
\cline{2-11} 
\multirow{1}{*}{USPST} & 99.5 & 100 & 99.67 & 99.50 & 99.40 & 99.44 & 100 & 99.61 & 99.44 & 99.44\tabularnewline
\cline{2-11} 
\multirow{1}{*}{AUSTRALIAN} & 87.18 & 86.67 & 87.07 & 86.38 & 86.06 & 85.47 & 84.80 & 85.51 & 84.09 & 84.09\tabularnewline
\cline{2-11} 
\multirow{1}{*}{A1A} & 83.75 & 83.08 & 82.89 & 83.13 & 83.10 & 61.25 & 59.06 & 58.99 & 59.10 & 59.10\tabularnewline
\cline{2-11} 
\multirow{1}{*}{MUSHROOMS} & 100 & 100 & 99.99 & 100 & 100 & 99.9 & 100 & 99.99 & 100 & 100\tabularnewline
\cline{2-11} 
\multirow{1}{*}{SVMGUIDE3} & 87.35 & 77.55 & 77.83 & 78.20 & 77.36 & 78.01 & 20.06 & 22.25 & 22.81 & 22.81\tabularnewline
\cline{2-11} 
\multirow{1}{*}{W5A } & 91.128 & 88.40 & 88.11 & 87.29 & 87.92 & 86.02 & 81.50 & 81.24 & 79.30 & 79.30\tabularnewline
\cline{2-11} 
\multirow{1}{*}{W8A} & 97.29 & 97.29 & 97.27 & 97.25 & 97.29 & 11.46 & 11.62 & 10.81 & 10.33 & 10.33\tabularnewline
\cline{2-11} 
\multirow{1}{*}{COD-RNA} & 97.39 & 97.42 & 97.36 & 97.40 & 97.36 & 22.61 & 21.15 & 20.23 & 22.41 & 22.41\tabularnewline
\cline{2-11} 
\multirow{1}{*}{IJCNN1} & 89.59 & 87.44 & 87.20 & 87.22 & 88.52 & 80.87 & 79.74 & 79.53 & 79.56 & 79.56\tabularnewline
\cline{2-11} 
\multirow{1}{*}{COVTYPE} & 87.71 & 80.99 & 81.03 & 80.90 & 80.79 & 89.05 & 85.63 & 85.64 & 85.54 & 85.54\tabularnewline
\hline 
\end{tabular}}
\par\end{centering}
\caption{The classification performance w.r.t. different fraction of unlabeled
data.\label{tab:balanced90-1-1}}
\end{table}
\begin{figure}[t]
\noindent \begin{centering}
\includegraphics[width=0.47\textwidth]{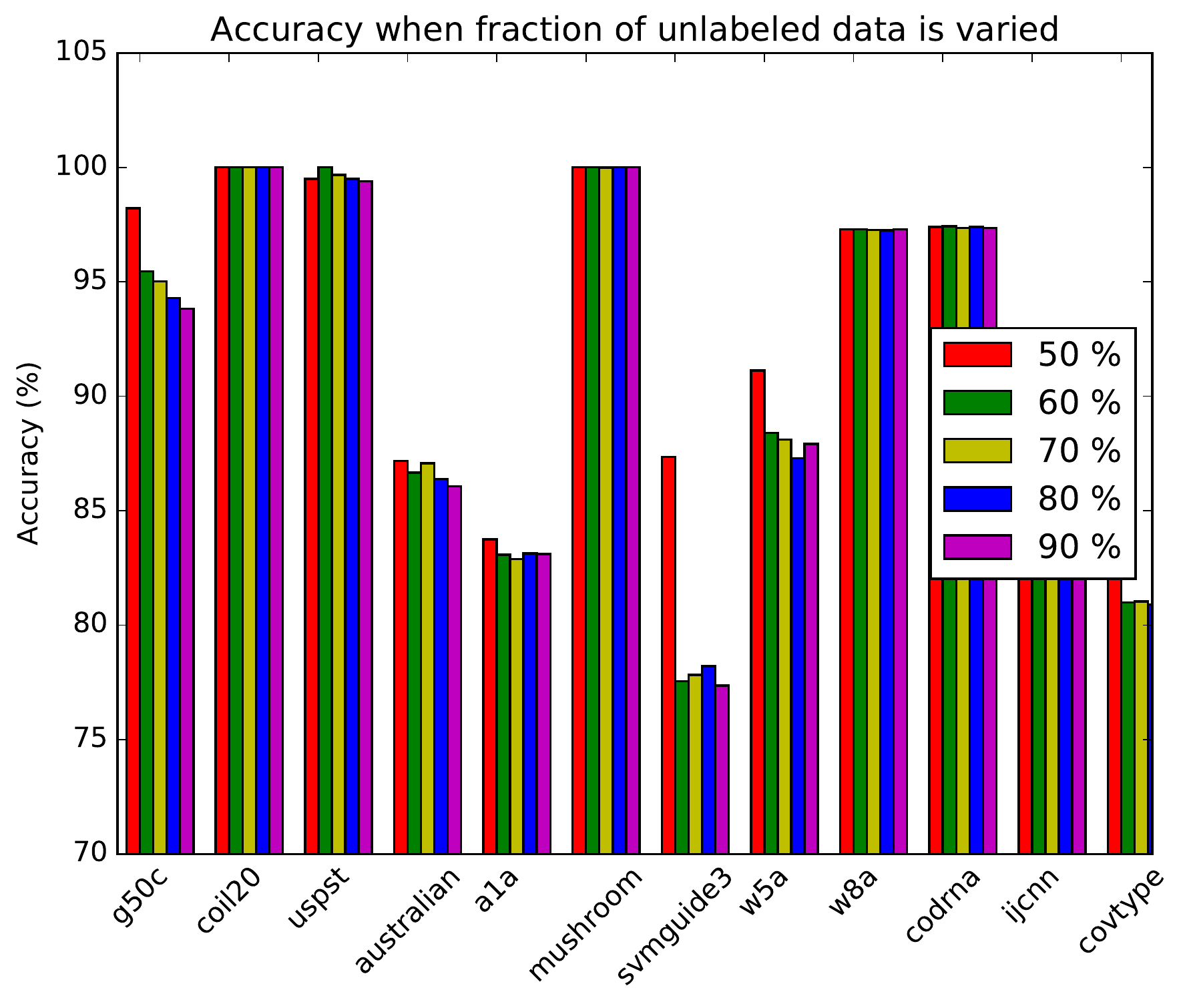}\includegraphics[width=0.47\textwidth]{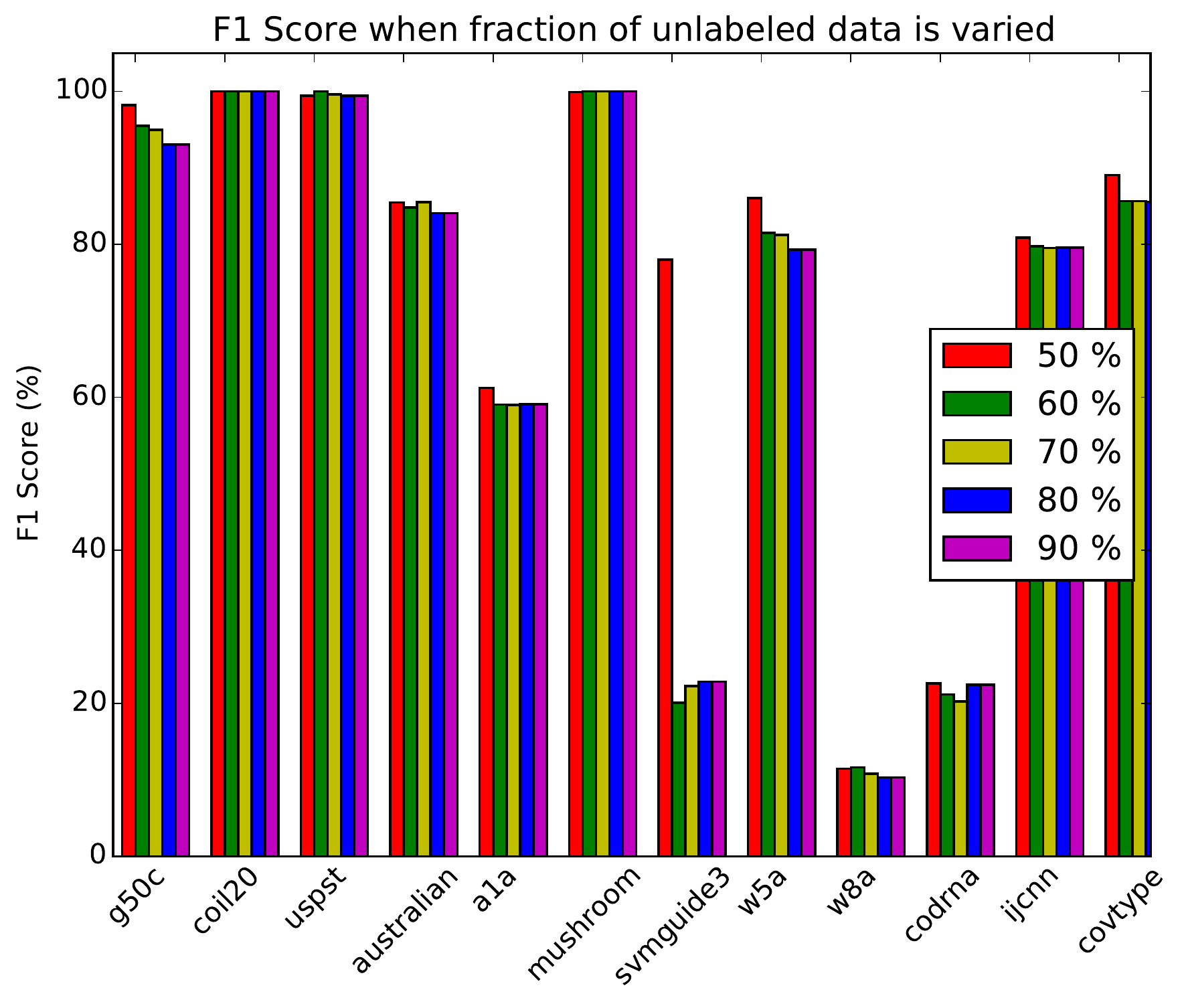}\vspace{-4mm}
\par\end{centering}
\caption{Left: The classification accuracy when fraction of unlabeled data
is varied. Right: The F1 score when fraction of unlabeled data is
varied.\label{fig:vary_unlabel}}
\end{figure}

In this simulation study, we address the question how the variation
in percentage of unlabeled data influences the learning performance.
We also experiment on the real datasets given in Table \ref{tab:detail}
with various proportions of unlabeled data varied in the grid $\left\{ 50\%,\,60\%,\,70\%,\,80\%,\,90\%\right\} $.
We observe that when the percentage of unlabeled data increases, the
classification accuracy and the F1 score tend to decrease across the
datasets except for two datasets COIL20 and MUSHROOMS which remain
fairly stable (cf. Table \ref{tab:balanced90-1-1} and Figure \ref{fig:vary_unlabel}
(left and right)). This observation is reasonable since when increasing
the percentage of hidden label, we decrease the amount of information
label provided to the classifier, hence making the label propagation
more challenging. \vspace{-2mm}

\subsection{Experimental Results on The Benchmark Datasets}

In this experiment, we compare our proposed method with LapSVM, CCCP,
Self-KNN and SVM as described in Section \ref{subsec:Baselines}.
Based on the best performance from the empirical model analysis in
Section \ref{subsec:exp2}, we use the combination of \emph{Hinge
loss} and the \emph{smoothness function} $l_{1}\left(.\right)$ for
our model. Besides offering the best predictive performance, this
combination also encourages the sparsity in the output solution. 

The RBF kernel, given by $K(x,x^{'})=\text{exp}(-\gamma||x-x^{'}||^{2})=\text{exp}(-\frac{1}{2\sigma_{l}^{2}}||x-x^{'}||^{2})$,
is used in the experiments. With LapSVM, we use the parameter settings
proposed in \citep{Melacci2011}, wherein the parameters $\gamma_{A}$
and $\gamma_{I}$ are searched in the grid $\left\{ 10^{-6},\,10^{-4},\,10^{-2},\,10^{-1},\,1,\,10,\,100\right\} $.
In all experiments with the LapSVM, we make use of the preconditioned
conjugate gradient version, which seems more suitable for the LapSVM
optimization problem \citep{Melacci2011}. With CCCP-TSVM, we use
the setting $\text{CCCP-TSVM}|_{UC*=LC}^{s=0}$. Only two parameters
need to be tuned are the trade-off $C$ and the width of kernel $\gamma$.
Akin to our proposed GKM, the trade-off parameters $C'=C$ is tuned
in the grid $\left\{ 2^{-5},2^{-3},\ldots,2^{3},2^{5}\right\} $ and
the width of kernel $\gamma$ is varied in the grid $\left\{ 2^{-5},\,2^{-3},\ldots,\,2^{3},\,2^{5}\right\} $.
In our proposed GKM, the bandwidth $\sigma_{s}$ of Gaussian kernel
weight function, which involves in computing the weights of the spectral
graph, is set to $\sigma_{s}=\sigma_{l}$. We split the experimental
datasets to $90\%$ for training set and $10\%$ for testing set and
run cross-validation with $5$ folds. The optimal parameter set is
selected according to the highest classification accuracy. We set
the number of iterations $T$ in GKM to $0.2\times\left(l+u\right)$
for the large-scale datasets such as MUSHROOMS, W5A, W8A, COD-RNA,
COVTYPE and IJCNN1 and to $l+u$ for the remaining datasets. Each
experiment is carried out five times to compute the average of the
reported measures.

\begin{table}
\begin{centering}
\resizebox{\textwidth}{!}{
\begin{tabular}{c|ccc|cc|c}
\hline 
\textbf{Dataset} & \textbf{Method} & \textbf{Acc 80\% } & \textbf{Acc 90\%} & \textbf{Time 80\% } & \textbf{Time 90\%} & \textbf{Memory} \tabularnewline
\hline 
\multirow{5}{*}{G50C} & GKM & 95.45 & \textbf{96.36} & \textbf{\textit{0.015}} & \textbf{\textit{0.015}} & 2.2\tabularnewline
 & LapSVM & 96.20 & 94.50 & 0.29 & 0.019 & 5.9\tabularnewline
 & CCCP & \textbf{98.18} & 94.55 & 0.14 & 0.509 & \uline{1.6}\tabularnewline
 & Self-KNN & 71.99 & 63.58 & 6.68 & 3.104 & 8.5\tabularnewline
 & SVM & 96.18 & 96.18 & 0.11 & 0.106 & 2.7\tabularnewline
\hline 
\multirow{5}{*}{COIL20} & GKM & \textbf{100.00} & \textbf{100.00} & \textbf{\textit{0.01}} & \textbf{\textit{0.005}} & 2.7\tabularnewline
 & LapSVM & \textbf{100.00} & \textbf{100.00} & 0.39 & 0.016 & 11\tabularnewline
 & CCCP & 98.10 & \textbf{100.00} & 1.1 & 0.366 & \uline{1.9}\tabularnewline
 & Self-KNN & 93.18 & 98.67 & 7.97 & 1.529 & 3.3\tabularnewline
 & SVM & 100.00 & 100.00 & 0.1 & 0.090 & 2.9\tabularnewline
\hline 
\multirow{5}{*}{USPST} & GKM & 99.17 & \textbf{100.00} & \textbf{\textit{0.031}} & \textbf{\textit{0.031}} & 3.9\tabularnewline
 & LapSVM & 99.20 & 99.60 & 0.28 & 0.038 & 9.3\tabularnewline
 & CCCP & 99.58 & \textbf{100.00} & 0.61 & 2.170 & 3.6\tabularnewline
 & Self-KNN & \textbf{99.82} & 99.57 & 35.4 & 16.69 & \uline{3.3}\tabularnewline
 & SVM & 99.80 & 99.80 & 0.49 & 0.49 & 5\tabularnewline
\hline 
\multirow{5}{*}{AUSTRALIAN} & GKM & \textbf{91.30} & \textbf{90.10} & \textbf{\emph{0.021}} & \textbf{\textit{0.026}} & 3.9\tabularnewline
 & LapSVM & 85.90 & 86.20 & 0.94 & 0.032 & 4\tabularnewline
 & CCCP & 81.88 & 89.85 & \textit{\emph{0.04}} & 0.031 & 3.6\tabularnewline
 & Self-KNN & 84.31 & 83.76 & 4.97 & 2.305 & 11\tabularnewline
 & SVM & 87.64 & 87.64 & 0.027 & 0.027 & \uline{2.3}\tabularnewline
\hline 
\multirow{5}{*}{A1A} & GKM & \textbf{81.00} & \textbf{83.49} & \textbf{\textit{0.073}} & \textbf{\textit{0.078}} & 3.3\tabularnewline
 & LapSVM & 80.10 & 81.60 & 0.2 & 0.049 & 3\tabularnewline
 & CCCP & 79.75 & 82.37 & 0.95 & 0.047 & \uline{0.7}\tabularnewline
 & Self-KNN & 77.71 & 77.63 & 32.47 & 30.724 & 27\tabularnewline
 & SVM & 83.12 & 83.12 & 0.34 & 0.340 & 15\tabularnewline
\hline 
\multirow{5}{*}{MUSHROOMS} & GKM & 99.94 & \textbf{99.96} & \textbf{\textit{0.031}} & \textbf{\textit{0.037}} & 49\tabularnewline
 & LapSVM & 98.80 & 97.50 & 5.26 & 0.334 & 337\tabularnewline
 & CCCP & \textbf{100.00} & 99.96 & 28.1 & 8.820 & 177\tabularnewline
 & Self-KNN & 82.92 & 83.97 & 551 & 6,626 & \uline{37}\tabularnewline
 & SVM & 100.00 & 100.00 & 632 & 632.36 & 106\tabularnewline
\hline 
\multirow{5}{*}{SVMGUIDE3} & GKM & 80.24 & 81.05 & \textbf{\textit{0.015}} & \textbf{\textit{0.015}} & 1.5\tabularnewline
 & LapSVM & 75.80 & 77.90 & 0.33 & 0.028 & 2.9\tabularnewline
 & CCCP & 81.45 & \textbf{83.37} & 1.42 & 0.054 & \uline{0.7}\tabularnewline
 & Self-KNN & \textbf{88.24} & 91.28 & 18.37 & 19.21 & 10\tabularnewline
 & SVM & 83.67 & 83.67 & 0.23 & 0.23 & 6\tabularnewline
\hline 
\multirow{5}{*}{W5A } & GKM & 97.69 & \textbf{97.61} & \textbf{\textit{0.041}} & \textbf{\textit{0.052}} & 103\tabularnewline
 & LapSVM & 97.00 & 97.50 & 1.18 & 0.521 & 811\tabularnewline
 & CCCP & \textbf{98.33} & 97.39 & 146.28 & 7.41 & 251\tabularnewline
 & Self-KNN & 77.50 & 65.34 & 1,778 & 1,001 & \uline{84}\tabularnewline
 & SVM & 98.49 & 98.49 & 48.1 & 48.06 & 106\tabularnewline
\hline 
\multirow{5}{*}{W8A} & GKM & \textbf{97.42} & \textbf{98.00} & \textbf{\textit{0.073}} & \textbf{\textit{0.677}} & \uline{110}\tabularnewline
 & LapSVM & 97.40 & 97.32 & 26.15 & 9.150 & 17,550\tabularnewline
 & CCCP & 97.10 & 97.18 & 1,380 & 379.06 & 277\tabularnewline
 & Self-KNN & 73.87 & 71.06 & 38,481 & 27,502 & 388\tabularnewline
 & SVM & 98.82 & 98.82 & 64.65 & 64.65 & 116\tabularnewline
\hline 
\multirow{5}{*}{COD-RNA} & GKM & 87.43 & 87.51 & \textbf{\textit{0.995}} & \textbf{\textit{1.115}} & \uline{110}\tabularnewline
 & LapSVM & 85.70 & 86.10 & 13.15 & 11.42 & 12,652\tabularnewline
 & CCCP & \textbf{88.48} & \textbf{89.74} & 3,900 & 326.72 & 279\tabularnewline
 & Self-KNN & 61.43 & 63.26 & 31,370 & 27,568 & 649\tabularnewline
 & SVM & 92.04 & 92.04 & 7,223 & 7,223 & 117\tabularnewline
\hline 
\multirow{5}{*}{IJCNN1} & GKM & 93.12  & \textbf{92.62} & \textbf{\textit{0.672 }} & \textbf{\textit{0.729 }} & \uline{111}\tabularnewline
 & LapSVM & \textbf{95.30} & 80.90 & 15.4 & 8.08 & 17,015\tabularnewline
 & CCCP & 93.09 & 93.29 & 6,718 & 302.81 & 274.3\tabularnewline
 & Self-KNN & 92.93 & 91.97 & 12,988 & 12,739 & 336\tabularnewline
 & SVM & 98.84 & 98.84 & 301 & 301.48 & 117\tabularnewline
\hline 
\multirow{5}{*}{COVTYPE} & GKM & 83.89 & 85.29 & \textbf{\textit{1.42}} & \textbf{\textit{1.51}} & \uline{119.5}\tabularnewline
 & LapSVM & 81.80 & 80.20 & 19.8 & 34.02 & 69,498\tabularnewline
 & CCCP & \textbf{85.91} & \textbf{85.75} & 5,958 & 1,275 & 292\tabularnewline
 & Self-KNN & 64.30 & 70.02 & 19,731 & 47,875 & 892\tabularnewline
 & SVM & 90.06 & 90.06 & 3,535 & 3,535 & 130\tabularnewline
\hline 
\end{tabular}}
\par\end{centering}
\caption{The classification accuracy (\%), training time (Time) (second), and
used memory (MB of RAM) of the competitive methods when $80\%$ and
$90\%$ of data are hidden label.\label{tab:balanced80}}
\end{table}

We measure the accuracy, training time, and memory amount used in
training when the percentages of unlabeled data are $80\%$ and $90\%$.
These measures are reported in Table \ref{tab:balanced80}. To improve
the readability, in these two tables we emphasize the best method
(not count the full-labeled SVM) for each measure using boldface,
italicizing, or underlining. Regarding the classification accuracy,
it can be seen that GKM are comparable with LapSVM and CCCP while
being much better than Self-KNN. Particularly, CCCP seems to be the
best method on $80\%$ unlabeled dataset  while GKM slightly outperforms
others on $90\%$ unlabeled dataset.  Comparing with the full-labeled
SVM, except for three datasets IJCNN1, COD-RN, and COVTYPE, GKM produces
the comparable classification accuracies. Remarkably for the computational
time, GKM outperforms the baselines by a wide margin especially on
the large-scale datasets.  On the large-scale datasets W8A, COD-RNA,
IJCNN1, and COVTYPE, GKM is significantly tens of times faster than
LapSVM, the second fastest method. We also examine the memory consumption
in training for each method. It can be observed that GKM is also economic
in terms of memory amount used in training especially on the large-scale
datasets.  Our GKM consistently uses the least amount of memory in
comparison to other methods especially on the large-scale datasets.
In contrast, LapSVM always consumes a huge amount of memory during
its training. In summary, our GKM is promising to be used in real-world
applications since it is scalable, accurate, and economic in memory
usage. Most importantly, GKM is the first online learning method for
kernelized semi-supervised learning.

\vspace{-3mm}

\section{Conclusion\label{sec:conclusion}}

In this paper, we present a novel framework for semi-supervised learning,
called Graph-based Semi-supervised Kernel Machine (GKM). Our framework
conjoins three domains of kernel method, spectral graph, and stochastic
gradient descent. The proposed GKM can be solved directly in the primal
form with the ideal convergence rate $\text{O}\left(\frac{1}{T}\right)$
and naturally inherits all strengths of an SGD-based method. We validate
and compare GKM with other state-of-the-art methods in semi-supervised
learning on several benchmark datasets. The experimental results demonstrate
that our proposed GKM offers comparable classification accuracy and
is efficient in memory usage whilst achieving a significant speed-up
comparing with the state-of-the-art baselines. Moreover, our approach
is the first semi-supervised model offering the online setting that
is essential in many real-world applications in the era of big data.

\clearpage{}

\appendix

\section{Appendix on Convex Analysis \label{sec:Convex-Analysis}}

\noindent \textbf{Proof of Lemma \ref{lem:w_t}}

\noindent 
\begin{align}
\overline{\bw}_{i} & =\frac{i-2}{i}\overline{\bw}_{i-1}+\frac{2}{i}\bw_{i-1}\nonumber \\
i\left(i-1\right)\overline{\bw}_{i} & =\left(i-1\right)\left(i-2\right)\overline{\bw}_{i-1}+2\left(i-1\right)\bw_{i-1}\label{eq:lem1_1}
\end{align}
Taking sum Eq. (\ref{eq:lem1_1}) when $i=2,\ldots,t$, we gain
\begin{align*}
t\left(t-1\right)\overline{\bw}_{t} & =2\sum_{i=2}^{t}\left(i-1\right)\bw_{i-1}=2\sum_{i=1}^{t-1}i\bw_{i}
\end{align*}
\[
\overline{\bw}_{t}=2t^{-1}\left(t-1\right)^{-1}\sum_{i=1}^{t-1}i\bw_{i}
\]
\vspace{-2mm}

\noindent \textbf{Proof of Lemma \ref{lem:func}.} We consider three
cases as follows

\noindent i) In this case, we have $f\left(M;a,b,p\right)=aM^{p-1}-M+b=M^{p-1}\left(a-M^{2-p}\right)+b$.
Since $M=\max\left(1,\left(a+b\right)^{\frac{1}{2-p}}\right)$, we
have $M^{p-1}\geq1$ and $M^{2-p}\geq a+b>a$. Hence, we gain
\[
f\left(M;a,b,p\right)\leq a-M^{2-p}+b\leq0
\]

\noindent ii) With $p=2$ and $a<1$, we have $M=\frac{b}{1-a}>0$
and $f\left(M;p,a,b\right)=\left(a-1\right)M+b=0$

\noindent iii) In this case, we have
\begin{align*}
f\left(M;a,b,p\right) & =\frac{a}{\left(p-1\right)^{\frac{p-1}{p-2}}a^{\frac{p-1}{p-2}}}-\frac{1}{\left(p-1\right)^{\frac{1}{p-2}}a^{\frac{1}{p-2}}}+b\\
 & =\frac{a\left(\left(ab^{p-2}\left(p-1\right)^{p-1}\right)^{\frac{1}{p-2}}-\left(\left(p-2\right)^{p-2}\right)^{\frac{1}{p-2}}\right)}{\left(p-1\right)^{\frac{p-1}{p-2}}a^{\frac{p-1}{p-2}}}\leq0
\end{align*}
\vspace{-2mm}

\noindent 

\noindent \textbf{Proof of Lemma \ref{lem:wt_norm}. }We prove that
$\norm{\bw_{t}}\leq M$ for all $t$ by induction. Assume the hypothesis
holds with $t$, we verify it for $t+1$. We start with\vspace{-2mm}
\begin{align*}
\bw_{t+1} & =\bw_{t}-\eta_{t}g_{t}=\frac{t-1}{t+1}\bw_{t}-\frac{2C}{t+1}\nabla_{\bw}l\left(\bw_{t};x_{i_{t}},y_{i_{t}}\right)-\frac{2C'\mu_{u_{t}v_{t}}}{t+1}\nabla_{\bw}l_{p}\left(\transp{\bw_{t}}\Phi_{u_{t}v_{t}}\right)\\
\norm{\bw_{t+1}} & \leq\frac{t-1}{t+1}\norm{\bw_{t}}+\frac{2C}{t+1}\norm{\nabla_{\bw}l\left(\bw_{t};x_{i_{t}},y_{i_{t}}\right)}+\frac{2C'\mu_{u_{t}v_{t}}}{t+1}\norm{\nabla_{\bw}l_{p}\left(\transp{\bw_{t}}\Phi_{u_{t}v_{t}}\right)}\\
 & \leq\frac{t-1}{t+1}\norm{\bw_{t}}+\frac{2CA}{t+1}+\frac{2C'\mu_{u_{t}v_{t}}\norm{\nabla_{\bw}l_{p}\left(\transp{\bw_{t}}\Phi_{u_{t}v_{t}}\right)}}{t+1}\\
\norm{\nabla_{\bw}l_{p}\left(\transp{\bw_{t}}\Phi_{u_{t}v_{t}}\right)} & \leq\norm{p\text{sign}\left(\transp{\bw_{t}}\Phi_{u_{t}v_{t}}\right)\vert\transp{\bw_{t}}\Phi_{u_{t}v_{t}}\vert^{p-1}\Phi_{u_{t}v_{t}}}\\
 & \leq2Rp\norm{\Phi_{u_{t}v_{t}}}^{p-1}\norm{\bw_{t}}^{p-1}\leq\left(2R\right)^{p}p\norm{\bw_{t}}^{p-1}
\end{align*}

\noindent where we have $\norm{\Phi_{u_{t}v_{t}}}=\norm{\Phi\left(x_{u_{t}}\right)-\Phi\left(x_{v_{t}}\right)}\leq2R$.
Therefore, we gain the following inequality\vspace{-2mm}
\begin{align*}
\norm{\bw_{t+1}} & \leq\frac{t-1}{t+1}\norm{\bw_{t}}+\frac{2CA}{t+1}+\frac{2C'\mu_{u_{t}v_{t}}\left(2R\right)^{p}p\norm{\bw_{t}}^{p-1}}{t+1}\\
 & =\frac{t-1}{t+1}\norm{\bw_{t}}+\frac{2b}{t+1}+\frac{2a\norm{\bw_{t}}^{p-1}}{t+1}\,\left(\text{since}\,\,\ensuremath{\mu_{u_{t}v_{t}}\leq1}\right)
\end{align*}
where we denote $a=C'\left(2R\right)^{p}p$ and $b=CA$. Recall that
we define $M$ as
\[
M=\begin{cases}
\max\left(1,\left(a+b\right)^{\frac{1}{2-p}}\right) & \text{if}\,p<2\\
\frac{b}{1-a} & \text{if}\,p=2,a<1\\
\left(\frac{1}{\left(p-1\right)a}\right)^{\frac{1}{p-2}} & \text{if}\,p>2,ab^{p-2}\leq\frac{\left(p-2\right)^{p-2}}{\left(p-1\right)^{p-1}}
\end{cases}
\]
Referring to Lemma \ref{lem:func}, we have $f\left(M;p,a,b\right)\leq0$
and gain\vspace{-2mm}
\begin{align*}
\norm{\bw_{t+1}} & \leq\frac{t-1}{t+1}M+\frac{2b}{t+1}+\frac{2aM^{p-1}}{t+1}\\
 & \leq M+\frac{2\left(aM^{p-1}-M+b\right)}{t+1}\leq M+\frac{f\left(M;p,a,b\right)}{t+1}\leq M
\end{align*}
Therefore, the hypothesis holds for $t+1$. It concludes this proof.

\noindent \textbf{Proof of Lemma \ref{lem:gt}.} To bound $\norm{g_{t}}$,
we derive as\vspace{-1mm}
\begin{align*}
\norm{g_{t}} & \leq\norm{\bw_{t}}+C\norm{\nabla_{\bw}l\left(\bw_{t};x_{i_{t}},y_{i_{t}}\right)}+C'\mu_{u_{t}v_{t}}\norm{\nabla_{\bw}l_{p}\left(\transp{\bw_{t}}\Phi_{u_{t}v_{t}}\right)}\\
 & \leq M+CA+C'\left(2R\right)^{p}pM^{p-1}=G
\end{align*}
\vspace{-2mm}

\noindent \textbf{Proof of Theorem \ref{thm:regret}}{\small{}
\begin{align*}
\Vert\bw_{t+1}-\bw^{*}\Vert^{2} & =\Vert\bw_{t}-\eta_{t}g_{t}-\bw^{*}\Vert^{2}=\Vert\bw_{t}-\bw^{*}\Vert^{2}+\eta_{t}^{2}\Vert g_{t}\Vert^{2}-2\eta_{t}\transp{g_{t}}\left(\bw_{t}-\bw^{*}\right)
\end{align*}
}\vspace{-2mm}
{\small{}Taking the conditional expectation w.r.t $\bw_{t}$, we gain}\vspace{-2mm}

\noindent {\small{}
\begin{align*}
\mathbb{E}\left[\Vert\bw_{t+1}-\bw^{*}\Vert^{2}\right] & =\mathbb{E}\left[\Vert\bw_{t}-\bw^{*}\Vert^{2}\right]+\eta_{t}^{2}\mathbb{E}\left[\Vert g_{t}\Vert^{2}\right]-2\eta_{t}\mathbb{E}\left[\transp{g_{t}}\left(\bw_{t}-\bw^{*}\right)\right]\\
 & =\mathbb{E}\left[\Vert\bw_{t}-\bw^{*}\Vert^{2}\right]+\eta_{t}^{2}\mathbb{E}\left[\Vert g_{t}\Vert^{2}\right]+2\eta_{t}\transp{\left(\bw^{*}-\bw_{t}\right)}\nabla_{\bw}\mathcal{J}\left(\bw_{t}\right)\\
 & \leq\mathbb{E}\left[\Vert\bw_{t}-\bw^{*}\Vert^{2}\right]+\eta_{t}^{2}\mathbb{E}\left[\Vert g_{t}\Vert^{2}\right]+2\eta_{t}\left(\mathcal{J}\left(\bw^{*}\right)-\mathcal{J}\left(\bw_{t}\right)-\frac{1}{2}\Vert\bw_{t}-\bw^{*}\Vert^{2}\right)
\end{align*}
}{\small \par}

\noindent Taking expectation of the above equation again, we yield\vspace{-2mm}
{\small{}
\begin{align*}
\mathbb{E}\left[\Vert\bw_{t+1}-\bw^{*}\Vert^{2}\right] & \leq\mathbb{E}\left[\Vert\bw_{t}-\bw^{*}\Vert^{2}\right]+\eta_{t}^{2}\mathbb{E}\left[\Vert g_{t}\Vert^{2}\right]\\
 & +2\eta_{t}\left(\mathcal{J}\left(\bw^{*}\right)-\mathbb{E}\left[\mathcal{J}\left(\bw_{t}\right)\right]-\frac{1}{2}\mathbb{E}\left[\Vert\bw_{t}-\bw^{*}\Vert^{2}\right]\right)\\
 & \leq\left(1-\eta_{t}\right)\mathbb{E}\left[\Vert\bw_{t}-\bw^{*}\Vert^{2}\right]+G^{2}\eta_{t}^{2}+2\eta_{t}\left(\mathcal{J}\left(\bw^{*}\right)-\mathbb{E}\left[\mathcal{J}\left(\bw_{t}\right)\right]\right)\\
\mathbb{E}\left[\mathcal{J}\left(\bw_{t}\right)\right]-\mathcal{J}\left(\bw^{*}\right) & \leq\left(\frac{1}{2\eta_{t}}-\frac{1}{2}\right)\mathbb{E}\left[\Vert\bw_{t}-\bw^{*}\Vert^{2}\right]-\frac{1}{2\eta_{t}}\mathbb{E}\left[\Vert\bw_{t+1}-\bw^{*}\Vert^{2}\right]+\frac{G^{2}\eta_{t}}{2}
\end{align*}
}{\small \par}

\noindent Using the learning rate $\eta_{t}=\frac{2}{t+1}$, we gain\vspace{-2mm}
\begin{align*}
\mathbb{E}\left[t\mathcal{J}\left(\bw_{t}\right)\right]-t\mathcal{J}\left(\bw^{*}\right) & \le\frac{\left(t-1\right)t}{4}\mathbb{E}\left[\Vert\bw_{t}-\bw^{*}\Vert^{2}\right]-\frac{t\left(t+1\right)}{4}\mathbb{E}\left[\Vert\bw_{t+1}-\bw^{*}\Vert^{2}\right]+\frac{G^{2}t}{t+1}
\end{align*}

\noindent Taking sum when $t$ runs from $1$ to $T$, we achieve\vspace{-2mm}
\begin{align*}
\mathbb{E}\left[\sum_{t=1}^{T}t\mathcal{J}\left(\bw_{t}\right)\right]-\frac{T\left(T+1\right)}{2}\mathcal{J}\left(\bw^{*}\right) & \leq-\frac{T\left(T+1\right)}{4}\mathbb{E}\left[\bw_{T+1}-\bw^{*}\Vert^{2}\right]+\sum_{t=1}^{T}\frac{G^{2}t}{t+1}<G^{2}T\\
\mathbb{E}\left[\frac{2}{T\left(T+1\right)}\sum_{t=1}^{T}t\mathcal{J}\left(\bw_{t}\right)\right]-\mathcal{J}\left(\bw^{*}\right) & <\frac{2G^{2}}{T}\\
\mathbb{E}\left[\mathcal{J}\left(\frac{2}{T\left(T+1\right)}\sum_{t=1}^{T}t\bw_{t}\right)-\mathcal{J}\left(\bw^{*}\right)\right] & <\frac{2G^{2}}{T}\\
\mathbb{E}\left[\mathcal{J}\left(\overline{\bw}_{T+1}\right)-\mathcal{J}\left(\bw^{*}\right)\right] & <\frac{2G^{2}}{T}
\end{align*}

\noindent Furthermore, from the strong convexity of $\mathcal{J}\left(.\right)$
and $\bw^{*}$ is a minimizer, we have{\small{}
\begin{align*}
\mathcal{J}\left(\overline{\bw}_{T+1}\right)-\mathcal{J}\left(\bw^{*}\right) & +\frac{1}{2}\norm{\overline{\bw}_{T+1}-\bw^{*}}^{2}\geq\transp{\nabla_{\bw}\mathcal{J}\left(\bw^{*}\right)}\left(\overline{\bw}_{T+1}-\bw^{*}\right)\geq\frac{1}{2}\norm{\overline{\bw}_{T+1}-\bw^{*}}^{2}\\
\mathbb{E}\left[\norm{\overline{\bw}_{T+1}-\bw^{*}}^{2}\right] & <\frac{4G^{2}}{T}
\end{align*}
}\vspace{-3mm}

\noindent \textbf{Proof of Theorem} \textbf{\ref{thm:high_prob_J}.}
Let us denote the random variable $Z_{T+1}=\mathcal{J}\left(\overline{\bw}_{T+1}\right)-\mathcal{J}\left(\bw^{*}\right)\geq0$.
From Markov inequality, we have\vspace{-2mm}
\begin{align*}
\mathbb{P}\left[Z_{T+1}\geq\varepsilon\right] & \leq\frac{\mathbb{E}\left[Z_{T+1}\right]}{\varepsilon}=\frac{\mathcal{\mathbb{E}}\left[\mathcal{J}\left(\overline{\bw}_{T+1}\right)-\mathcal{J}\left(\bw^{*}\right)\right]}{\varepsilon}<\frac{2G^{2}}{T\varepsilon}\\
\mathbb{P}\left[Z_{T+1}<\varepsilon\right] & >1-\frac{2G^{2}}{T\varepsilon}
\end{align*}
\vspace{-2mm}
By choosing $T_{0}=\left\lceil \frac{2G^{2}}{\varepsilon\delta}\right\rceil $,
for all $T\geq T_{0}$, we have $\mathbb{P}\left[Z_{T+1}<\varepsilon\right]>1-\frac{2G^{2}}{T\varepsilon}\geq1-\delta$.

\bibliographystyle{plainnat}
\phantomsection\addcontentsline{toc}{section}{\refname}\bibliography{GKM}

\end{document}